\documentclass{article}

\usepackage{arxiv}

\usepackage[square]{natbib}
\usepackage{booktabs}
\usepackage{color}
\usepackage{makecell}
\usepackage{multirow}
\usepackage{algorithm}
\usepackage{algorithmicx}
\usepackage[noend]{algpseudocode}
\usepackage{amsmath}
\usepackage{amssymb}
\usepackage{bm}
\newtheorem{thm}{Theorem}
\newtheorem{prop}{Proposition}
\usepackage{subcaption}
\usepackage[export]{adjustbox}

\usepackage[utf8]{inputenc} 
\usepackage[T1]{fontenc}    
\usepackage{hyperref}       
\usepackage{url}            
\usepackage{booktabs}       
\usepackage{amsfonts}       
\usepackage{nicefrac}       
\usepackage{microtype}      
\usepackage{lipsum}

\title{LINDT: Tackling Negative Federated Learning with Local Adaptation}

\author{
  
  Hong Lin \\
  College of Computer Science and Technology\\
  Zhejiang University\\
  Zhejiang, China\\
  \texttt{honglin@zju.edu.cn} \\
   \And
  Lidan Shou\\
  College of Computer Science and Technology\\
  Zhejiang University\\
  Zhejiang, China\\
  \texttt{should@zju.edu.cn} \\
  \And
  Ke Chen\\
  College of Computer Science and Technology\\
  Zhejiang University\\
  Zhejiang, China\\
  \texttt{chenk@zju.edu.cn} \\
  \And
  Gang Chen\\
  College of Computer Science and Technology\\
  Zhejiang University\\
  Zhejiang, China\\
  \texttt{cg@zju.edu.cn} \\
  \And
  Sai Wu\\
  College of Computer Science and Technology\\
  Zhejiang University\\
  Zhejiang, China\\
  \texttt{wusai@zju.edu.cn} \\
}

\begin{document}
\maketitle

\begin{abstract}
Federated Learning (FL) is a promising distributed learning paradigm, which allows a number of data owners (also called clients) to collaboratively learn a shared model without disclosing each client's data. However, FL may fail to proceed properly, amid a state that we call negative federated learning (NFL). This paper addresses the problem of negative federated learning. We formulate a rigorous definition of NFL and analyze its essential cause. We propose a novel framework called LINDT for tackling NFL in run-time. The framework can potentially work with any neural-network-based FL systems for NFL detection and recovery. Specifically, we introduce a metric for detecting NFL from the server. On occasion of NFL recovery, the framework makes adaptation to the federated model on each client's local data by learning a Layer-wise Intertwined Dual-model. Experiment results show that the proposed approach can significantly improve the performance of FL on local data in various scenarios of NFL.
\end{abstract}


\section{Introduction} \label{intro}
 
Federated Learning is a promising distributed learning paradigm which allows a number of data owners (also called \textit{clients} or \textit{devices}) to collaboratively learn a shared model without disclosing their private data \citep{McMahan2017CommunicationEfficientLO}.
A typical FL system relies on iterative rounds of client-server interactions to find the global optimal model for all clients. Each round consists of three steps: (1) the server broadcasts the current global model to the participating clients; (2) 
each client updates the model independently to obtain a local optimum (i.e. a set of local parameters); and (3) the server collects the local optimums, then aggregates them to obtain an updated global model. 
This powerful paradigm has found a wide range of practical applications where data is decentralized and privacy is important \citep{Hard2018FederatedLF,Chen2019FedHealthAF}. 

Unfortunately, the success of FL is not always guaranteed.
It is known that the global model produced by FL may significantly under-perform centralized learning, and may even lose to models that clients train independently on their own \citep{Yu2020SalvagingFL,Smith2017FederatedML}. Failure of FL has been reported on real-world datasets from Reddit comments \citep{reddit}, tweets  \citep{Caldas2018LEAFAB}, Google glass \citep{7349385}, smartphone sensors \citep{Anguita2013APD}, vehicle sensors \citep{DUARTE2004826} etc. Thereby clients, especially those having rich private data for independent learning, can hardly benefit from FL and thus would have no incentive to participate in FL any more. 

The community has well recognized setback issues inherent in FL. To name a few, these include: (1) data distributions differ across clients; (2) system is under attack; (3) measures are taken to protect client privacy; and (4) clients become inactive unexpectedly.
Despite such empirical observations, no research has ever discussed common reasons for the failure of FL in various scenarios.
In addition, many questions remain open in the search for \textit{cure} of failed FL, for example: How to define and detect a failed FL process? 
Is it possible to achieve good local performance on clients when the global FL keeps failing?  And if yes what are the expenses?


In this paper, we attempt to answer the above questions. We coin a new term \textit{Negative Federated Learning (NFL)} to refer to the state of an FL system in which the iterative client-server interactions do not help clients in learning. We analyze various scenarios of NFL and find that significant divergence between the global optimum and local optimums is an essential cause to NFL. Therefore, by measuring the divergence, it is possible to detect NFL in the system in run-time. To tackle NFL, the system has to take recovery measures. Instead of trying to learn a better global model fitting wildly different client data distributions (and other issues), our recovery strategy aims at improving the performance of the federated model by making local adaptations on clients.

We summarize the contributions of our paper as follows:

1) We propose a novel framework called LINDT for addressing NFL in a federated learning system. The framework can be employed on any neural-network-based FL systems to \textit{detect} NFL and then to \textit{recover} from it. To the best of our knowledge, this is the first approach for tackling NFL in run-time.

2) We present rigorous proof for the use of our metric to detect NFL. We also propose the recovery method, which improves the performance of federated model by learning a \textit{layer-wise intertwined dual-model} on each client. 

3) We conduct extensive experiments on federated image classification and language modeling tasks. The results confirm the effectiveness of LINDT in detecting NFL and recovering from it.

\section{Related Work}\label{rw}

Federated learning is inherent with specific properties which make it markedly different from traditional distributed learning. These include, but not limited to: (1) \textit{Non-IID Data}. The data held by different clients tend to show non-independent-and-identical distributions \citep{Kairouz2019AdvancesAO}; (2) \textit{Vulnerable to Attacks.} FL systems are exposed to an environment where any participant including the central server can be unreliable. Potential attacks to FL, such as data poisoning \citep{Liu2018TrojaningAO} and model poisoning \citep{Bagdasaryan2020HowTB,Bhagoji2019AnalyzingFL}, are omnipresent;  (3) \textit{Privacy Protection.} Privacy is usually a key concern for the clients in FL. To ensure privacy, the system may take protective measures such as differential privacy on its broadcasted information \citep{McMahan2018LearningDP}; (4) \textit{Client Inactivity.} The clients in FL may become inactive unexpectedly, due to connection problem or hardware failure \citep{Wang2019FederatedEO}.
The clients may also fail to respond in time due to slow execution. 

Recent studies report that any of the above properties can pose negative effects on an FL system.
\citet{Zhao2018FederatedLW} observed the accuracy of a model learned by FL may drop over 50\% when data distributions differ across clients. \citet{Bhagoji2019AnalyzingFL} presented the model learned in an FL system could be easily manipulated by attackers to generate false predictions. \citet{Yu2020SalvagingFL} argued the differential privacy could also induce a significant performance drop in a federated model. 
Both \citet{McMahan2017CommunicationEfficientLO} and \citet{Briggs2020FederatedLW} showed that client inactivity would harm the convergence of an FL-trained model. Many real-world FL tasks reportedly suffer from a mixture of all these negative effects, and thus fail to outperform independent learning on clients \citet{Yu2020SalvagingFL,Smith2017FederatedML}.

The above observations motivate the proposal of many solutions against the negative impact on FL, such as sharing a public dataset to all clients for balancing their different data distributions \citep{Zhao2018FederatedLW}, designing robust aggregation against attacks \citep{pmlr-v80-yin18a}, fine-tuning an FL-trained model on individual clients \citep{Wang2019FederatedEO,Mansour2020ThreeAF}, as well as utilizing the technique of user clustering \citep{Sattler2019ClusteredFL}, meta learning \citep{Jiang2019ImprovingFL} or multi-task learning \citep{Smith2017FederatedML} to augment FL. These studies do provide useful ideas for analyzing and addressing what we call Negative Federated Learning (NFL). However, these studies are inadequate, as they lack rigorous analysis of the essential cause of NFL and dynamic solution to it.

%
We consider NFL as a dynamic behavior that may occur in any FL systems. The problem is measurable, detectable and avoidable, and that when a system enters NFL, it can utilize our proposed approach to recover to a healthy state. Our approach handles NFL mainly by making local adaptation on clients. 
In view of similar approaches in  \citep{Hanzely2020FederatedLO,Deng2020AdaptivePF}, however, our approach LINDT has the following advantages: (1) Our dual-model training is more flexible due to its per-layer mixing strategy. (2) LINDT introduces an instance-wise weighting scheme for better
fitting data samples.


\section{Negative Federated Learning}
In this section, we present the definition of NFL and analyze its cause. We shall start by introducing federated learning.

\label{sec3} 
\subsection{Preliminaries and Notations}
Given a system of $N$ clients, each of which has its own private dataset $D_i$ (where $i \in \{1,\ldots,N\}$), and is willing to learn a shared global model, then the conventional \textbf{centralized learning} approach requires all private data to be pooled together for training such a model. However, due to privacy concerns and limitation in network bandwidth, this approach is impractical and is used for reference only.


\setlength{\tabcolsep}{3pt}
\begin{table}[h]
  \centering
  \small
    \begin{tabular}{ll}
    \toprule
    $(x, y)$ & a data sample and its label in $D_{i}$ \\
    $n_i$ & number of data samples $(x, y) \in D_{i}$ \\
    $n$ & number of data samples $(x, y) \in \cup_{i=1}^{N} D_i$ \\
    $Y$ & a set consisting of all unique values of $y$ in $\cup_{i=1}^{N} D_i$ \\
    $G$ & a model trained in federated learning system \\
    $\boldsymbol{w}$ & weight vector (also called \textit{parameter set}) of $G$\\ 
    \bottomrule
    \end{tabular}%
  
  \caption{Main Notations.}
  \label{notation}%
\end{table}%


\textbf{Federated Learning (FL)} is a learning process, in which $N$ clients collaboratively train a model $G(\cdot, \boldsymbol{w})$ without the necessity for any data owner $i$ to expose its data $D_{i}$ to other clients $j$, where $j \in \{1,\ldots,N\}$ and $j \neq i$. Let $p_i$ be the distribution of the data samples in $D_{i}$,
and $\ell (\boldsymbol{w}, x, y)$ be the loss of the prediction on data sample $(x,y)$, we can formulate the FL problem into a form with the following objective:
\begin{equation}
\boldsymbol{w}^{*} = \arg \min \limits_{\boldsymbol{w}}  \sum_{i=1}^{N}  \frac{n_i}{n} \mathbb{E}_{(x,y) \sim p_i} \ell (\boldsymbol{w}, x, y).
\end{equation}


The optimal parameter set $\boldsymbol{w}^{*}$ is typically obtained via an iterative process of client-server interactions, as shown in Algorithm \ref{fedavg_algorithm}, where after the last round $r$, $\boldsymbol{w}^{r}$ is taken as $\boldsymbol{w}^{*}$. Detailed descriptions of the algorithm are omitted. 

Ideally, the performance of model $G(\cdot,\boldsymbol{w}^{r})$ improves by $r$ and approaches that of the imaginary centralized learning\footnote{Unless otherwise stated, superscripts on vectors mean the number of iterative rounds throughout the entire paper.}. In such case, the FL algorithm is said to \textit{converge}. Obviously, the main incentive for the clients to continue contributing to FL is to obtain better parameters from the server (Parameter set $\boldsymbol{w}$ is \textit{better} than $\boldsymbol{w}'$ if its respective model $G(\cdot,\boldsymbol{w})$ outperforms $G(\cdot,\boldsymbol{w}')$ on a certain dataset).
Therefore, clients participating in an FL process expect $\boldsymbol{w}^{r}$ to be better than their local optimums $\boldsymbol{w}_i^{r}$ which are trained on their private data $D_i$. However, it turns out that FL can go wrong when such expectation is not met.

\renewcommand{\algorithmicrequire}{ \textbf{Server executes:}} 
\renewcommand{\algorithmicensure}{ \textbf{Clients execute:}} 

\begin{algorithm}[t] 
\small
\caption{FedAvg \citep{McMahan2017CommunicationEfficientLO}}
\label{fedavg_algorithm} 

\textbf{Input:} A set of clients $i \in \{1,\ldots,N\}$. The number of clients that perform computation in each round, $K$. Local mini-batch size, $B$. The number of local epochs, $E$. Learning rate, $\eta$.

\algrenewcommand{\alglinenumber}[1]{s\footnotesize#1:}%
\begin{algorithmic}[1] 
\Require 
\State initialize global model with weights $\boldsymbol{w}^{0}$
\For{each round $r$=1,2,\ldots}
    \State broadcast the latest weight vectors $\boldsymbol{w}^{r-1}$ to all clients
    \State wait until receiving $K$ locally-updated weight vectors $\boldsymbol{w}^{r}_{i}$ 
    \Statex \hspace{\algorithmicindent}from a set of active clients $C_r$
    \State \label{s5} obtain a new \textit{global optimum}  via $\boldsymbol{w}^{r}\leftarrow\frac{1}{K}\sum_{i \in C_r} \boldsymbol{w}^{r}_{i}$ \footnotemark
\EndFor
\end{algorithmic}

\algrenewcommand{\alglinenumber}[1]{c\footnotesize#1:}%
\begin{algorithmic}[1]
\Ensure 
\For{each round $r$=1,2,\ldots}
    \State download the latest weight vectors $\boldsymbol{w}^{r-1}$ from server
    \State $\boldsymbol{w}_{i}^{r} \leftarrow \boldsymbol{w}^{r-1}$
    \State $\mathcal{B}_i \leftarrow$ split local data $D_i$ into batches of size $B$
    \For{each local epoch from 1 to $E$}
        \For{batch  $b_i \in \mathcal{B}_i$}
            \State  \label{step12} $\boldsymbol{w}_{i}^{r} \leftarrow \boldsymbol{w}_{i}^{r} - \eta \nabla_{\boldsymbol{w}_{i}^{r}} \mathbb{E}_{(x,y) \in b_i} \ell (\boldsymbol{w}_{i}^{r}, x,y)$
        \EndFor
    \EndFor
    \State \label{step13} upload the \textit{local optimum} $\boldsymbol{w}_{i}^{r}$ to server
\EndFor
\end{algorithmic}
\end{algorithm}
\footnotetext{The original FedAvg used \textit{weighted mean} to aggregate $\boldsymbol{w}_i^r$. For clarity in presentation we simply use average instead.}

\subsection{Definition of Negative Federated Learning}

To decide if FL is doing well, let us \textit{pretend} that every client $i$ has a private model ($P_i$) trained independently on $D_i$ beforehand.
\textit{Negative Federated Learning (NFL)} refers to the state of an FL system in which the model obtained from FL does not win out over $P_i$.

To formalize the definition of NFL, we first define a metric describing the performance gain obtained from FL by a client $i$. Given an FL-trained model $G$ and a private model $P_i$
, we denote by $V^{G}_{i}$ the performance\footnote{Most existing performance metrics, such as accuracy or F1, can be used for this purpose.} of $G$, and by $V^{P}_{i}$ the performance of $P_i$, the \textit{on-client performance gain} is given by
\begin{equation}\label{positive2}
\abovedisplayskip
\belowdisplayskip
    \beta_i = V^{G}_{i} - V^{P}_{i}.
\end{equation}
Note that in order to compute meaningful performance gain, each local model is required to have the \textit{same} neural structure as the global model.

Next, we define a system-wide metric $\beta$, named \textit{weighted performance gain}, as following: 
\begin{equation}\label{positive3}
\abovedisplayskip
\belowdisplayskip
    \beta = \sum_{i=1}^{N} \alpha_{i} \beta_i
\end{equation}
where weight $\alpha_i$ indicates how much client $i$ matters in the performance gain evaluation. Possible weight schemes could be $\alpha_i = \frac{1}{N}$ (equal weights), $\alpha_i = \frac{n_i}{n}$ (weight by client data size), or a positive value indicating the quality of local data \citep{inbook}.

Now we define NFL as follows: For a given federated learning system (i.e., running Algorithm 1), if there does NOT exist a positive integer $R$~ s.t. for any model $G(\cdot, \boldsymbol{w}^{r})$ learned  in this system after round $R$ (i.e. $r \geq R$), $\beta \geq 0$, then we say the system is in \textit{negative federated learning}. In such case, $|\beta|$ can be used to quantify the negative effects on the participating clients. 

\begin{figure}[t]
\centering
\includegraphics[width=0.7\columnwidth]{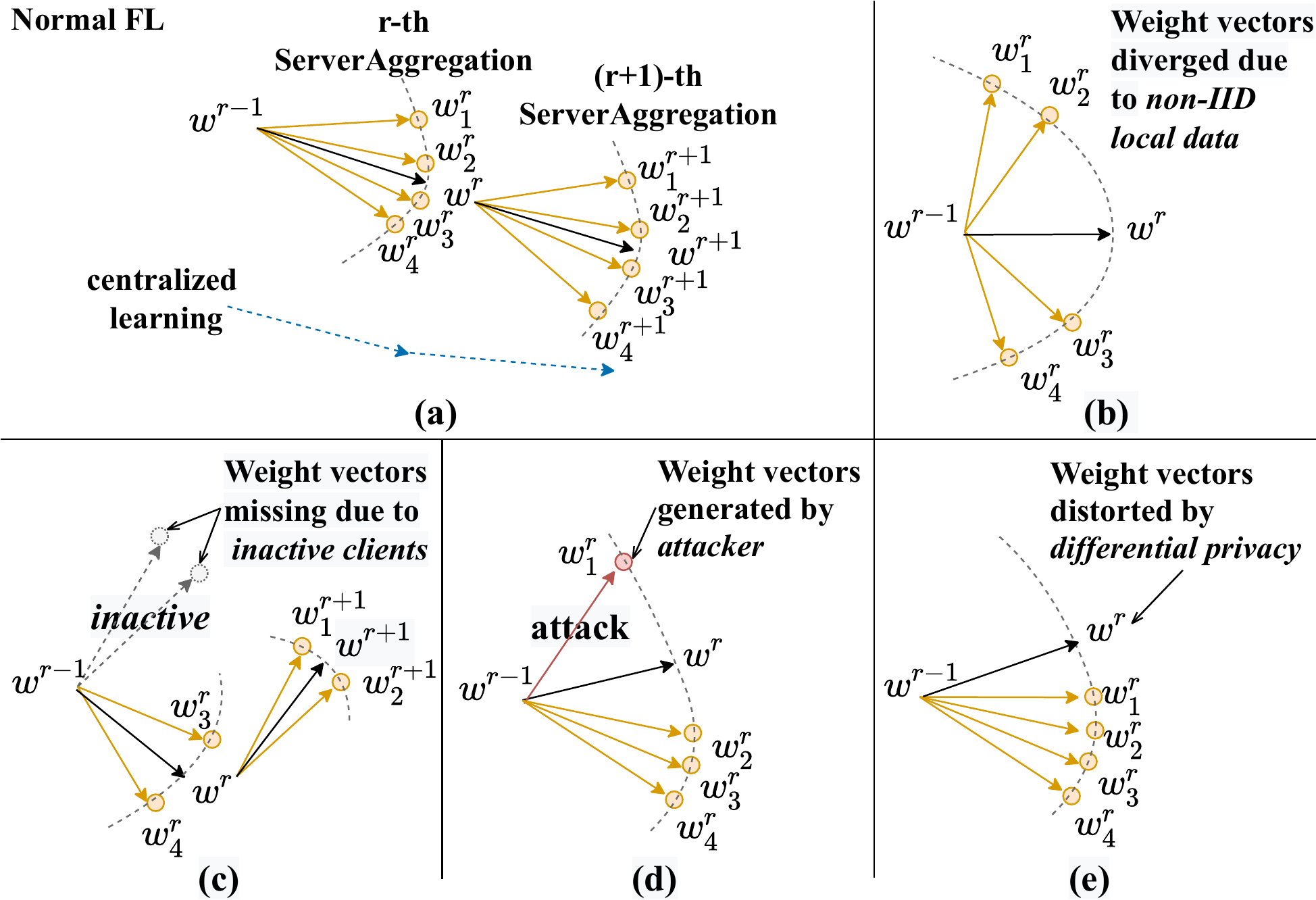}
\caption{2D plot of parameter updating of 4 clients. (a) normal FL, (b-e) negative  learning cases. \textit{Orange arrows:} local update traces. \textit{Black arrows:} global parameter update trace. \textit{Dot blue arrows:} a reference trace for centralized learning. }
\label{nfl}
\end{figure}

\subsection{Causes of Negative Federated Learning}\label{3.3}
While there is no general agreement on the inherent system properties which cause NFL, a careful study of issues leading to NFL can reveal its root cause.
We illustrate these issues by plotting in 2D the traces of updates to the global/local parameter sets ($\boldsymbol{w}$/$\boldsymbol{w}_i$), as shown in Fig.~\ref{nfl}.


Fig.~\ref{nfl}(a) shows the update trace of the global parameter set $\boldsymbol{w}$ (black arrows) and four local update traces $\boldsymbol{w}_i$ (orange arrows) in two rounds of a normal FL process.  It can be seen that the global trace moves steadily closer towards the blue trace which indicates the trace of centralized learning.

Fig.~\ref{nfl}(b,c,d,e) show four different cases of problematic updates. Fig.~\ref{nfl}(b) shows a common case of differing data distributions among clients, where four client traces diverge drastically, making the update in global trace unpredictable. 

Fig.~\ref{nfl}(c) shows another common issue where client 1 and 2 become inactive during the $r$-th iteration, while client 3 and 4 become inactive during ($r+1$)-th iteration. The global update trace turns violently, due to the missing local updates. 

Fig.~\ref{nfl}(d) illustrates a case where an attacker launches model poisoning attack on client 1. It deliberately reports an abnormal local update to manipulate the global trace. Such attack causes high error rate indiscriminately for all clients.

Fig.~\ref{nfl}(e) presents a case in which the system applies differential privacy protection on weight vectors to prevent disclosure of client privacy \citep{McMahan2018LearningDP}. 
The system typically replaces line s\ref{s5} in Algorithm 1 with the following steps: (1) clip the norm of the local updates with an upper bound $S$; and (2) add Gaussian noise, which also introduces distortion in global parameters: 
\begin{equation}\label{dp}
    \boldsymbol{w}^{r} = \boldsymbol{w}^{r-1} + \frac{1}{K} \sum_{i \in C_r} Clip(\boldsymbol{w}_{i}^{r} - \boldsymbol{w}^{r-1} , S) + \mathcal{N}(\boldsymbol{0}, \sigma^2 \boldsymbol{I}).
\end{equation}

Case (b) through (e) are similar in that the  
global optimum $\boldsymbol{w}^r$ is diverged significantly from the local optimums $\boldsymbol{w}^r_i$. As a result, a global model $G(\cdot,\boldsymbol{w}^{r})$ is highly probable to perform poorly on individual clients. These observations, though inexhaustive, all lead to a tentative proposition that \textit{the large divergence between the global optimum and local optimums could be an essential cause of NFL.} 

\subsection{Detecting Negative Federated Learning} \label{wdiv}
Basically, NFL can be detected by checking the value of $\beta$ after each round of interaction. However, on-client testing of the FL-trained model in each round is costly, as it incurs extra computing and transmission on clients. Inspire by \citep{Zhao2018FederatedLW},
we advocate a server-based method using the divergence in global/local optimums as mentioned in the previous section.

Specifically, given weight vectors $\boldsymbol{w}_{i}^{r}$ reported from the $K$ active clients $i \in C_r$ and their aggregated results $\boldsymbol{w}^{r}$, the weight divergence $w\_div^{r}$ for the $r$-th round of interaction is defined as 
\begin{equation}\label{wdivr}
    w\_div^{r} = \frac{1}{K} \sum_{i \in C_r} 
    \Vert \boldsymbol{w}_{i}^{r} - \boldsymbol{w}^{r}\Vert.
\end{equation}

A large $w\_div^{r}$ means significant differences between the local optimums $\boldsymbol{w}_{i}^{r}$ and the global optimum $\boldsymbol{w}^{r}$, and implies the likelihood of NFL! 
An advantage of $w\_div^{r}$ is that it is very easy to compute on the server once the aggregation for the $r$-th round is finished.

We now present a theoretical bound for $w\_div^{r}$ which is crucial for NFL detection.

\begin{prop}
     Given an FL system (i.e., running Algorithm \ref{fedavg_algorithm} and optionally applying differential privacy as described in Eq.~\ref{dp}), suppose each client $i$ $(i \in \{1,\ldots,N\} )$ has the following properties: 1) has $n_i$ samples following its own data distribution $p_i$, 2) makes $T$ steps of mini-batch SGDs in each round of execution, and 3) $\nabla_{\boldsymbol{w}} \mathbb{E}_{x|y } \ell (\boldsymbol{w}, x, y)$ is $\lambda_{x|y }$-Lipschitz for each label $y \in Y$, then for any integer $r>0$,
    \begin{equation} \label{inequity}
    \begin{aligned}
    w\_div^{r} &\leq  \Vert \mathcal{N}(\boldsymbol{0}, \sigma^2 \boldsymbol{I})\Vert  + \frac{\eta}{K^2} \sum_{i \in C_r} \sum_{j \in C_r \atop j \neq i}\\
    & \quad  \sum_{y \in Y}  \Vert p_i(y)-  p_j(y) \Vert  \sum_{t=0}^{T-1} (a_i)^{t} g_{max}(\boldsymbol{w}_{j}^{(r,T-1 -t)}).
    \end{aligned}
    \end{equation}
    where $a_i = 1+\eta \sum_{y \in Y} \lambda_{x|y } p_i(y)$ and $g_{max}(\boldsymbol{w}) = \max_{y \in Y}\Vert\nabla_{\boldsymbol{w}} \mathbb{E}_{x|y } \ell (\boldsymbol{w}, x, y)\Vert$. Note that $r$ and $T-1-t$ in the parentheses superscripts above indicate round and mini-batch, but $t$ in $(a_i)^t$ means exponent.
\label{prop1}
\end{prop}

Detailed proof of Proposition  \ref{prop1} can be found in Appendix \ref{A1}. This inequality reveals that $w\_div^{r}$ is bounded by two terms. The first is introduced by the standard deviation (std, $\sigma$) used to generate Gaussian noise in the optional differential privacy, while the second is mainly determined by the number of responded clients $K$, the differences in client data distributions $ \Vert p_i(y)-  p_j(y) \Vert $, and the gradients calculated on each clients $g_{max}(\boldsymbol{w}_{j}^{(r,T-1-t)})$. In fact, the second term should approach zero in a converging FL process as $r$ increments. Therefore, we can derive the following theorem:

\begin{thm}\label{thm1}
If the gradients $g_{max}(\boldsymbol{w}_{j}^{(r,T-1-t)})$ calculated on each client $j$ (where $j \in \{1,\ldots,N\}$),  all converge to zero as $r \rightarrow +\infty$, then, for any $\epsilon >0 $, there exists an integer $r' >0$ such that when $r>r'$,
\begin{equation}\label{eq:theorem}
    \Delta= w\_div^r - \Vert \mathcal{N}(\boldsymbol{0}, \sigma^2 \boldsymbol{I}))\Vert \leq \epsilon.
\end{equation}

\end{thm}

Note both $w\_div^r$ and $\Vert \mathcal{N}(\boldsymbol{0}, \sigma^2 \boldsymbol{I})\Vert $ are maintained on the server, so $\Delta$ can be easily evaluated by the server.
According to Theorem \ref{thm1}, 
if $\Delta$ does not approach zero after a certain number of rounds, we conclude that the FL process does not converge and thus the system is entering NFL.

\subsubsection{NFL Detection} Based on above discussions, we can develop an NFL detection scheme on the server as following:
\begin{itemize}
    \item Before starting FL, we predefine two system parameters, a threshold $\epsilon > 0$ and a sufficiently large integer $r'$.
    \item During iterative FL, the server evaluates if $ \Delta > \epsilon $ after each round of client-server interactions. If it is true in more than $r'$ rounds, the server reports a state of NFL. 
\end{itemize}

Once NFL is detected in a system, the global model is not expected to perform well on individual clients. However, it is possible to make local adaptation to the global model for better \textit{local performance}.
In the next section, we propose a framework for making local adaptation on clients during federated learning.

\section{Dual-model Training}

To tackle NFL, our approach introduces for each client a \textit{local model} ($L_i(\cdot, \boldsymbol{v}_i)$), which can fit the local data better and is tightly coupled in the learning process of the \textit{global model}. This process, called \textit{dual-model training}, aims at mitigating the negative effects on the client by optimizing the performance of the dual-model on the local data. 

\begin{figure}[t]
\centering
\includegraphics[width=0.7\columnwidth]{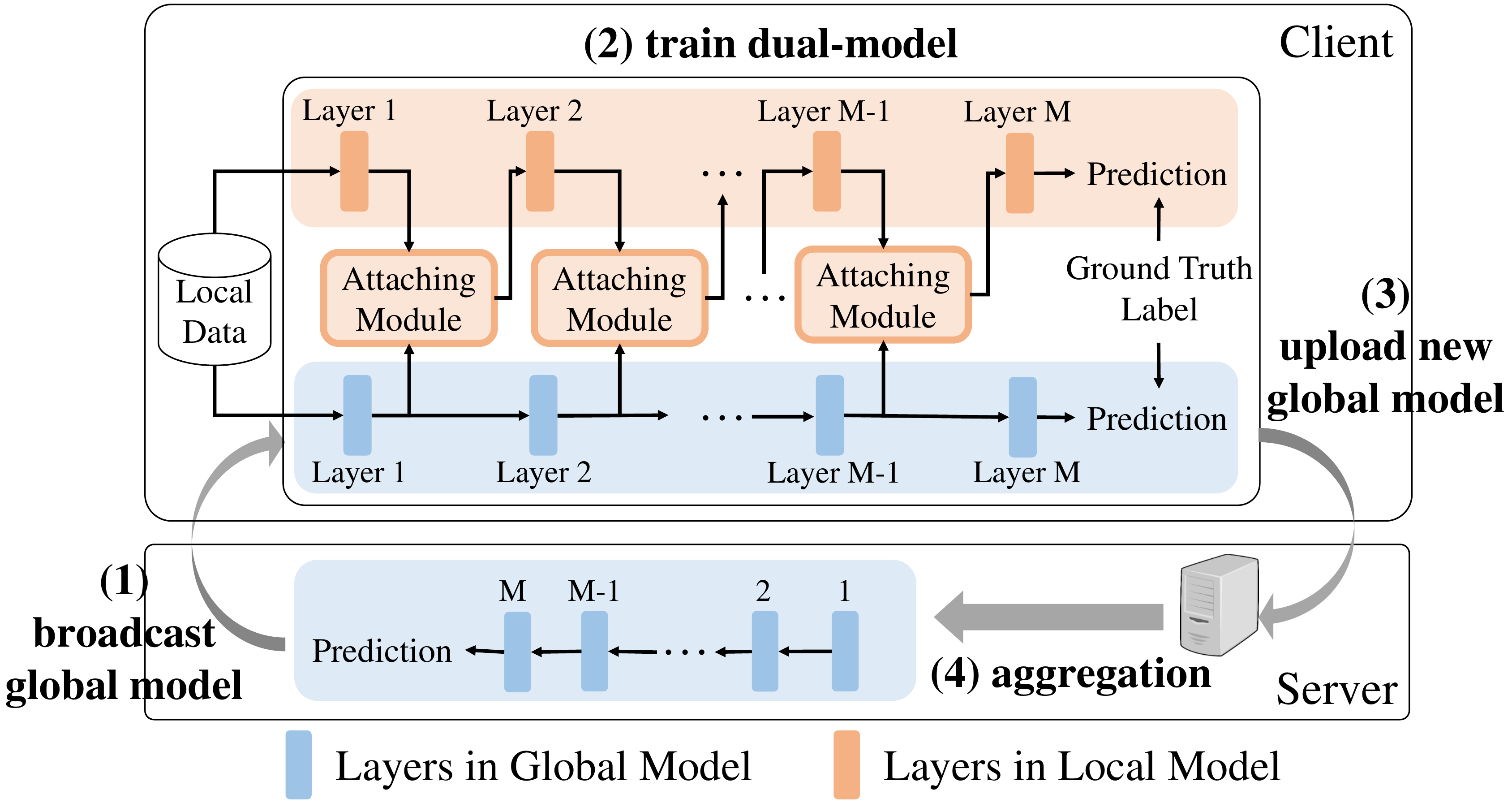}
\caption{An overview of Layer-wise INtertwined Dual-model Training (LINDT). During \textit{training}, a client leverages the prediction from both global and local models to calculate the loss. But when \textit{testing}, each client only considers the prediction from its local model (top orange box).} 
\label{overview}
\end{figure}

As Fig.~\ref{overview} shows, the global model and the local one are intertwined by a series of \textit{Attaching Modules} in the client. Layer $m$ of the local model takes input from the \mbox{($m-1$)-th} attaching module, which integrates the outputs from the global/local model at layer $m-1$. Such neural structure helps to enrich and generalize the information encoded in the per-layer hidden representation of the local model, thereby enhancing the generalization capability of the local model.

In the following texts, we focus on (1) the layer-wise intertwinement between the global and local models, and (2) the updating of the dual-model.

\subsection{Layer-wise Intertwinement} \label{sec:intertwinement}
Layer-wise intertwinement is implemented with a series of attaching modules. With these modules, the input to a layer of the local model integrates the outputs of both the global model and the local model from the previous layer.

Given a total of $M$ layers in the dual-model, we denote by $G^{m}(x, \boldsymbol{w}_{i}^{r})$ the representation of a data sample $x$ from the $m$-th layer of global model, and by $L_{i}^{m}(x, \boldsymbol{v}_i)$ the representation of the same data sample from the $m$-th layer of local model, then the intertwined representation $\boldsymbol{h}$ to be fed into the ($m+1$)-th layer of the local model is obtained by \textit{attaching} $G^{m}(x, \boldsymbol{w}_{i}^{r})$ to $L_{i}^{m}(x, \boldsymbol{v}_i)$, as given in the following equations:
\begin{equation}\label{attention1}
    score = sigmoid(\frac{ L_{i}^{m}(x, \boldsymbol{v}_i) \cdot  {G^{m}(x, \boldsymbol{w}_{i}^{r})}}{||L_{i}^{m}(x, \boldsymbol{v}_i)||})
\end{equation}
\begin{equation}\label{attention2}
    \boldsymbol{h} = score \cdot G^{m}(x, \boldsymbol{w}_{i}^{r}) + (1 - score) \cdot L_{i}^{m}(x, \boldsymbol{v}_i)
\end{equation}

$score$ in Eq.~\ref{attention1} quantifies the scalar projection of vector $G^{m}(x, \boldsymbol{w}_{i}^{r})$ in the direction of vector $L_{i}^{m}(x, \boldsymbol{v}_i)$, which is similar to scaled dot-product attention in \citep{NIPS2017_7181}.
A small $score$ indicates a large angle between $L_{i}^{m}(x, \boldsymbol{v}_i)$ and $G^{m}(x, \boldsymbol{w}_{i}^{r})$, which implies a contradiction at layer $m$ between the global and local representations of data $x$. For such data sample, Eq.~\ref{attention2} allows a loose intertwinement between the two models at layer $m$. This helps to prevent the local model from being contaminated by incompatible global information. Oppositely, if $score$ is large, there is more agreement between the two models at layer $m$, thus we can put more emphasis on the global representation. Such mechanism can help to enhance the local model with general features encoded in the global model.

\subsection{Dual-model Updates and LINDT Operations}
For updating the dual-model, we denote by $G^{M}(x, \boldsymbol{w}_{i}^{r})$ and $L_{i}^{M}(x, \boldsymbol{v}_{i})$, the respective pre-softmax logit outputs from the last layer of the dual-model (one \textit{global} and one \textit{local}), and let $\ell_{cross}$ be cross entropy. Each client $i$ objects to minimize
\begin{equation} \label{objective}
\begin{aligned}
    & \ell (\boldsymbol{w}_{i}^{r}, \boldsymbol{v}_{i}) = \mathbb{E}_{(x,y) \in D_i} [
    \ell_{cross}(G^{M}(x, \boldsymbol{w}_{i}^{r}), y) + 
    \ell_{cross}(L_{i}^{M}(x, \boldsymbol{v}_{i}), y) ].
\end{aligned}
\end{equation}

The training procedure of LINDT with \textit{NFL detection} can be easily implemented by making  amendments to Algorithm \ref{fedavg_algorithm}. Specifically, the operation of NFL detection is added on the server code after line s\ref{s5}. Besides, the training of the dual-model on the client is realized by replacing line c\ref{step12} in Algorithm \ref{fedavg_algorithm} with $\boldsymbol{w}_{i}^{r} \leftarrow \boldsymbol{w}_{i}^{r} - \eta \nabla_{\boldsymbol{w}_{i}^{r}} \ell (\boldsymbol{w}_{i}^{r}, \boldsymbol{v}_{i})$ and $\boldsymbol{v}_{i} \leftarrow \boldsymbol{v}_{i} - \eta \nabla_{\boldsymbol{v}_{i}} \ell (\boldsymbol{w}_{i}^{r}, \boldsymbol{v}_{i})$. The complete pseudo code of LINDT is shown in Algorithm \ref{alg:Framwork}, which is given in Appendix \ref{A2} due to space limit.

At the end of each round $r$ of dual-model training, clients are required to return the updated $\boldsymbol{w}_{i}^{r}$ to the server for aggregation (same as line c\ref{s5} in Algorithm \ref{fedavg_algorithm}). The server, upon receiving the first $K$ updates from the active clients ($C_r$), aggregates the parameters in the same way as conventional FL. Note that our design does not incur additional data exchange between the server and the clients.

There are two modes to use LINDT: (1) One is to use it for \textit{detection and recovery}, as described in Algorithm \ref{alg:Framwork}. An FL system running in this mode can operate normally without incurring dual-model training in the clients until it detects NFL. Once NFL occurs, a $flag$ is set and the clients start recovery by activating LINDT. The subsequent training and testing on the clients then go through LINDT. 
As NFL detection in the server is inexpensive, this mode is cost-effective. 
(2) Alternatively, the system may choose to always turn $flag$ on to use LINDT
in the entire learning life cycle. This mode would certainly incur more computation on the clients.

\section{Experimental Results}

We evaluate the performance of LINDT and previous approaches on two deep-learning tasks, namely CIFAR (\textit{image classification} on CIFAR-10 \citep{Krizhevsky2009LearningML}) and SHAKE (\textit{language modeling} on Shakespeare dataset \citep{Caldas2018LEAFAB}). Our test environment simulates pathologically negative effects by considering all the four cases as shown in Fig.~\ref{nfl}(b-e). Table \ref{task_profile} summarizes the task profiles and their default environment settings. Implementation details and extra experiment results are given in Appendix \ref{experiment_appedix}.

\setlength{\tabcolsep}{3pt}
\begin{table}[h]
  \centering
  \small
  \begin{tabular}{cccc}
    \toprule
     Task Name    & CIFAR & SHAKE \\
    \midrule
    Dataset & CIFAR-10 & Shakespeare \\
    
    \# Clients (\# Classes) & 100 (10) & 66 (80) \\
    Neural model & CNN   & LSTM \\    
    \multirow{2}{*}{Default data  allocation } & \multirow{2}{*}{non-IID(\textit{Mixed})}  & Each speaking \\
    &   & role as a client \\

    
    Per-round \% of active clients  & 10\% & 10\% \\ 
    
    Per-round \% of attackers & 20\% & 16\% \\
    
    $\mathcal{S}$ and $\sigma$  in diff. privacy & 15 (0.001)    & 15 (0.001) \\


    \bottomrule
    \end{tabular}%
  \caption{Task profiles.}
  \label{task_profile}%
\end{table}%

We mainly monitor three performance metrics: (1) the \textit{central accuracy}, evaluated for the global federated model on the server by pooling all test data together; (2) the average \textit{local accuracy}, evaluated for the local federated model (the dual-model for LINDT) on each client by testing independently on its local test data; and (3) the average performance gain $\beta$, given the private models pre-trained by clients independently on their own  \footnote{$\alpha_i=\frac{1}{N}$ in Eq. \ref{positive3}. 
Other weights result in the same conclusions.}. 

Each FL process is given fixed rounds of client-server interactions (i.e., $500$ rounds for CIFAR, $100$ rounds for SHAKE).
The average results over the last 10 rounds are reported. All experiments are repeated for 3 runs \footnote{Code can be found at ***Anonymous URL***.}.

\setlength{\tabcolsep}{5pt}
\begin{table*}[t]
  \centering
  \small %
    \begin{tabular}{lccccccccccccc}
    \toprule
    \makecell[l]{Task} & \multicolumn{3}{c}{CIFAR \textit{/NFL}} 
    &\multicolumn{3}{c}{SHAKE \textit{/NFL}} & 
    &\multicolumn{3}{c}{CIFAR \textit{/Normal FL}} 
    &\multicolumn{3}{c}{SHAKE \textit{/Normal FL}} \\
    \cmidrule(l{4pt}r{4pt}){2-4} \cmidrule(l{4pt}r{4pt}){5-7} \cmidrule(l{4pt}r{4pt}){9-11}\cmidrule(l{4pt}r{4pt}){12-14}
    
    \makecell[l]{Metric (accuracy)} & Central   & Local  & $\beta$ 
    & Central   & Local  & $\beta$ & 
    & Central   & Local  & $\beta$
    & Central   & Local  & $\beta$\\
    \midrule
    \textit{FedAvg}  & 55.63 & 56.19 & -17.83 
    & 43.60  & 44.26 & -2.93 &
    & 73.00 & 73.04 & + 29.99 & 52.93 & 53.42 & +6.16 \\
    \textit{FedProx} & 55.20 & 55.92 & -18.10 
    & 40.20 & 40.64 &  -6.55 &
    & 72.33 & 72.91 & + 29.86 &  52.07 & 52.62 & +5.36\\
    \textit{TrimmedMean}  & 55.67 & 56.70 & -17.32
    & 43.93  & 45.53 & -1.66 &
    & 71.47 & 72.09 & + 29.04 & 52.67 & 53.19 & +5.93 \\
    \textit{FB}  & -  & 81.17 & +7.15 
    & -   &  51.85 & +4.66 &
    & - & 61.19 & +18.14 & - & 53.89 & +6.63 \\
    \textit{APFL} & 54.90  & 80.61 & +6.59 
    & 43.87  & 51.96  & +4.77 &
     & 70.73 & 70.85 & +27.80 & 52.97 & 53.23 & +5.97\\
    \midrule          
    \textit{LINDT} & 55.73 & 81.61 & +7.59 
    & 49.17  & 52.84 & +5.65 & 
    &\textbf{73.07} & \textbf{73.30} & \textbf{+30.05}& \textbf{53.67} & \textbf{54.07} & \textbf{+6.81}\\
    \textit{LINDT-TM} & \textbf{56.23} & \textbf{82.82} &\textbf{+8.80} 
    & \textbf{49.86}  & \textbf{53.02} & \textbf{+5.83} &
    & - & - & - &- & - &-\\
   
    \bottomrule
    \end{tabular}
    \caption{Results of performance comparison with previous FL methods. Tasks tagged \textit{NFL} run in the default NFL environment as specified in Table \ref{task_profile}. The \textit{Normal FL} tasks get rid of most negative effects (CIFAR uses IID allocations whereas SHAKE remains unchanged; all attackers removed; default differential privacy and active rate). The central accuracy of FB is omitted as it is same as FedAvg. 
    } 
    
  \label{compare_others}%
\end{table*}%

\subsection{Comparison with Previous Methods} \label{compare_previous}
We compare LINDT with the following approaches: (1) the conventional FL approach \textit{FedAvg} \citep{McMahan2017CommunicationEfficientLO}, (2) \textit{FedProx} \citep{Sahu2018FederatedOF}, an approach proposed to tackle device and statistical heterogeneity in federated environment, (3) \textit{TrimmedMean} \citep{pmlr-v80-yin18a}, an approach that prevents the FL-trained global model from being poisoned, (4) \textit{FB} \citep{Wang2019FederatedEO}, a personalization method to fine-tune the top layer of the FL-trained model on the client, and (5) \textit{APFL} \citep{Deng2020AdaptivePF}, a recent work also integrating the global model with a per-client local model. 
For fair comparison, LINDT turns on dual-model training all the time.

\subsubsection{Main results}
Table \ref{compare_others} shows the results of all approaches in \textit{NFL} and \textit{normal FL} conditions. 
As the left half of the table shows, the first three methods, FedAvg, FedProx and TrimmedMean, end up in NFL. However, the rest three, FB, APFL, and LINDT, can handle the negative issues well, with LINDT leading in all three metrics. The results confirm the effectiveness of LINDT in tackling NFL.
Specifically, the gain in local accuracy of LINDT is significant in both datasets,
whereas its gain in central accuracy is much higher in SHAKE than in CIFAR. The last finding can be explained as follows: (1) SHAKE is under heavier attack (due to the nature of attack on language model). As a result, the other methods suffer heavier loss in central accuracy in NFL. In contrast, LINDT is much more resistant to heavy attacks. 
(2) The data distributions of SHAKE turn out to be more similar (e.g. same class labels across all clients). Therefore, LINDT manages to contribute better global parameter updates by dual-model training. 

The right half of Table \ref{compare_others} shows the results under normal FL environment. Surprisingly, most previous methods perform no better, if not worse, than FedAvg in normal condition. In contrast, LINDT can still perform well in all metrics on both datasets. Although the gaps are small, LINDT outperforms the other methods. We consider this feature an unexpected plus. 

Another revelation from Table \ref{compare_others} is that, although LINDT has been designed to optimize the local performance, it turns out to lead in central accuracy as well.  This is another advantage of LINDT, owing to the dual-model training.

\subsubsection{Integrating other techniques} 
As a flexible framework, LINDT can be further enhanced by integrating other FL techniques. A simple case is to replace the conventional aggregation method with the \textit{TrimmedMean}
aggregation \citep{pmlr-v80-yin18a} in the server. The respective results are shown in Table \ref{compare_others} as \textit{LINDT-TM}. It can be seen that such integration can further improve performance on all metrics at time of NFL. However, the integration is unnecessary in normal FL, since \textit{TrimmedMean} does not improve performance.

\subsection{Tuning Environment Settings}
\label{sec:tuning}

In this experiment, we tune the system environment parameters to study their impact on the performance of LINDT. We also analyze the results of \textit{FedAvg} for reference. In each test, we vary one parameter only, while keeping the rest values as default. Parameters with asterisk (*) indicate the default values in our experiment. Due to space limit, we present only the results on CIFAR, as those on SHAKE display similar trends/patterns and are given in Appendix \ref{experiment_appedix}.

\setlength{\tabcolsep}{3pt}

\begin{table}[t]
  \centering
  \small %
    \begin{tabular}{l|cc|cc|cc}
    \toprule
    \makecell[c]{Data} & \multicolumn{2}{c}{Central ACC} \vline  & \multicolumn{2}{c}{Local ACC} \vline & \multicolumn{2}{c}{$\beta$}\\
    \makecell[c]{Alloc.} & FedAvg & LINDT  & FedAvg & LINDT & FedAvg & LINDT \\
    \midrule
    
    IID & 60.30 & \textbf{60.43} & 60.75 & \textbf{69.87}   & +16.99 & \textbf{+26.10}\\
    
    non-IID & & & & \\
    
    \quad 10  & 55.30 & \textbf{55.40} & 56.44 & \textbf{ 76.77} &  -8.79 & \textbf{+11.54} \\
    
    \quad ~5 & 48.97& \textbf{49.40} &  50.64 & \textbf{84.06} & -25.85 & \textbf{+7.58} \\
    
    \quad ~2  & 32.50 & \textbf{33.23} &  32.19 & \textbf{93.73} & -58.12 & \textbf{+3.32} \\
    
    \textit{Mixed}*  & 55.63  & \textbf{55.73} & 56.19 & \textbf{81.61} & -17.83 & \textbf{+7.59}\\ 
    \bottomrule
    \end{tabular}%
    \caption{Varying data distributions among clients.}
    
  \label{noniid}%
\end{table}%

\begin{table}[t]
  \centering
  \small %
    \begin{tabular}{c|cc|cc|cc}
    \toprule
    \multirow{2}{*}{$K/N$} & \multicolumn{2}{c}{Central ACC} \vline & \multicolumn{2}{c}{Local ACC} \vline &  \multicolumn{2}{c}{$\beta$}  \\
     & FedAvg & LINDT  & FedAvg & LINDT  & FedAvg & LINDT \\
    \midrule
    90\%  & 57.37  & \textbf{57.87} & 58.14 & \textbf{82.89} & -15.88 & \textbf{+8.87}\\ 
         
    70\%  & 57.35  & \textbf{57.85} & 57.67 & \textbf{82.82} & -16.35 & \textbf{+8.80}\\ 
         
    50\%  & 57.19  & \textbf{57.47} & 57.76 & \textbf{81.89} & -16.26 & \textbf{+7.87}\\ 
         
    30\%  & 56.43 & \textbf{56.94} & 57.28 & \textbf{82.67} & -16.74 & \textbf{+8.65}\\ 
         
    10\%*   & 55.63  & \textbf{55.73} & 56.19 & \textbf{81.61} & -17.83 & \textbf{+7.59}\\ 
    \bottomrule
    \end{tabular}%
    \caption{Varying the ratio of active clients in each round.
    }

  \label{active}%
\end{table}%

\begin{table}[t]
  \centering
  \small 
    \begin{tabular}{c|cc|cc|cc}
    \toprule
    \multirow{2}{*}{Attack} & \multicolumn{2}{c}{Central ACC} \vline & \multicolumn{2}{c}{Local ACC} \vline &  \multicolumn{2}{c}{$\beta$}  \\
     & FedAvg & LINDT  & FedAvg & LINDT  & FedAvg & LINDT \\
    \midrule
    0\%  & 64.33  & \textbf{65.20} & 64.95 & \textbf{83.06} & -7.69 & \textbf{+10.42}\\ 
    
    10\%  & 56.87  & \textbf{57.90} & 57.80 & \textbf{81.76} & -15.30 & \textbf{+8.66}\\ 
         
    20\%*   & 55.63  & \textbf{55.73} & 56.19 & \textbf{81.61} & -17.83 & \textbf{+7.59}\\ 
         
    \bottomrule
    \end{tabular}%
    \caption{Varying the proportion of attackers in each round.}

  \label{attack}%
\end{table}%

\begin{table}[t]
  \centering
  \small 
    \begin{tabular}{c|cc|cc|cc}
    \toprule
    \multirow{2}{*}{$\sigma$} & \multicolumn{2}{c}{Central ACC} \vline & \multicolumn{2}{c}{Local ACC} \vline &  \multicolumn{2}{c}{$\beta$}  \\
     & FedAvg & LINDT  & FedAvg & LINDT  & FedAvg & LINDT \\
    \midrule
    0.001*   & 55.63  & \textbf{55.73} & 56.19 & \textbf{81.61} & -17.83 & \textbf{+7.59}\\ 
    
    0.003  & 52.35 & \textbf{52.69} & 52.61 & \textbf{80.63} & -21.41 & \textbf{+6.61}\\ 
    
    0.005  & 47.88 & \textbf{48.32} & 48.56 & \textbf{78.73} & -25.46 & \textbf{+4.71}\\ 
         
    0.007  &  40.84 & \textbf{42.12} & 41.09 & \textbf{76.17} & -32.93 & \textbf{+2.15}\\ 
    
    0.01  & 30.14  & \textbf{31.19} & 30.60 & \textbf{75.37} & -43.42 & \textbf{+1.35}\\ 
    \bottomrule
    \end{tabular}%
    \caption{Varying the \textit{std} ($\sigma$) of noises for differential privacy. }
  \label{sigma}%
\end{table}%

The tuning results are shown in Table \ref{noniid} to \ref{sigma}. Generally, NFL is prevalent in the entire parameter space that we tune.
The global federated model, in cases of NFL, can hardly reach high accuracy. In contrast, LINDT can always ensure a positive gain in local accuracy ($\beta>0$), and in many cases $\beta >5$.
Such gain is important, as it is a main reason for clients to participate in FL.

Table \ref{noniid} shows that data distributions do have major impact on the results. When the distributions become more different across clients \footnote{Number of classes allocated to each client varies from 10 to 2, thus increasing \textit{non-IIDness}. Details are given in Appendix \ref{experiment_appedix}.}, a major reduction in central and local ACC is evident for FedAvg. However, LINDT produces very high local ACC and slightly better central ACC compared to FedAvg. These results reveal that LINDT can make good local adaptation on the client data. Note that as the number of classes on each client reduces, independent learning performs much better (results not shown here). This explains the reduction in $\beta$.

Table \ref{active} demonstrates that client inactivity do pose some negative impact on FL, since both methods see reduction in central and local ACC when the ratio of active clients in each round varies from 90\% to 10\%. Nevertheless, the reduction is not significant, and the local performance gain achieved by LINDT is very stable (around 8 for all ratios).


Table \ref{attack} and Table \ref{sigma}
show the negative impact by variable poison attacks and noises introduced by differential privacy. As both negative issues increase quantitatively, the accuracy metrics of FedAvg reduce significantly. However, LINDT is much more resilient against these negative impact. In particular, LINDT provides a positive gain in local ACC ($\beta=1.35$) even when the central ACC of the global model reduces by over 24. The results further confirm the effectiveness of LINDT in well adapting to local data and thereby salvaging the local performance of FL on clients.

\subsection{Results of NFL Detection and Recovery} \label{sec:detect_recovery}



\begin{figure}[htbp]
\centering
\makebox[0pt]{
    \hskip -0.05cm
     \begin{subfigure}[b]{0.49\columnwidth}
         \includegraphics[width=\columnwidth,height=0.75\columnwidth]{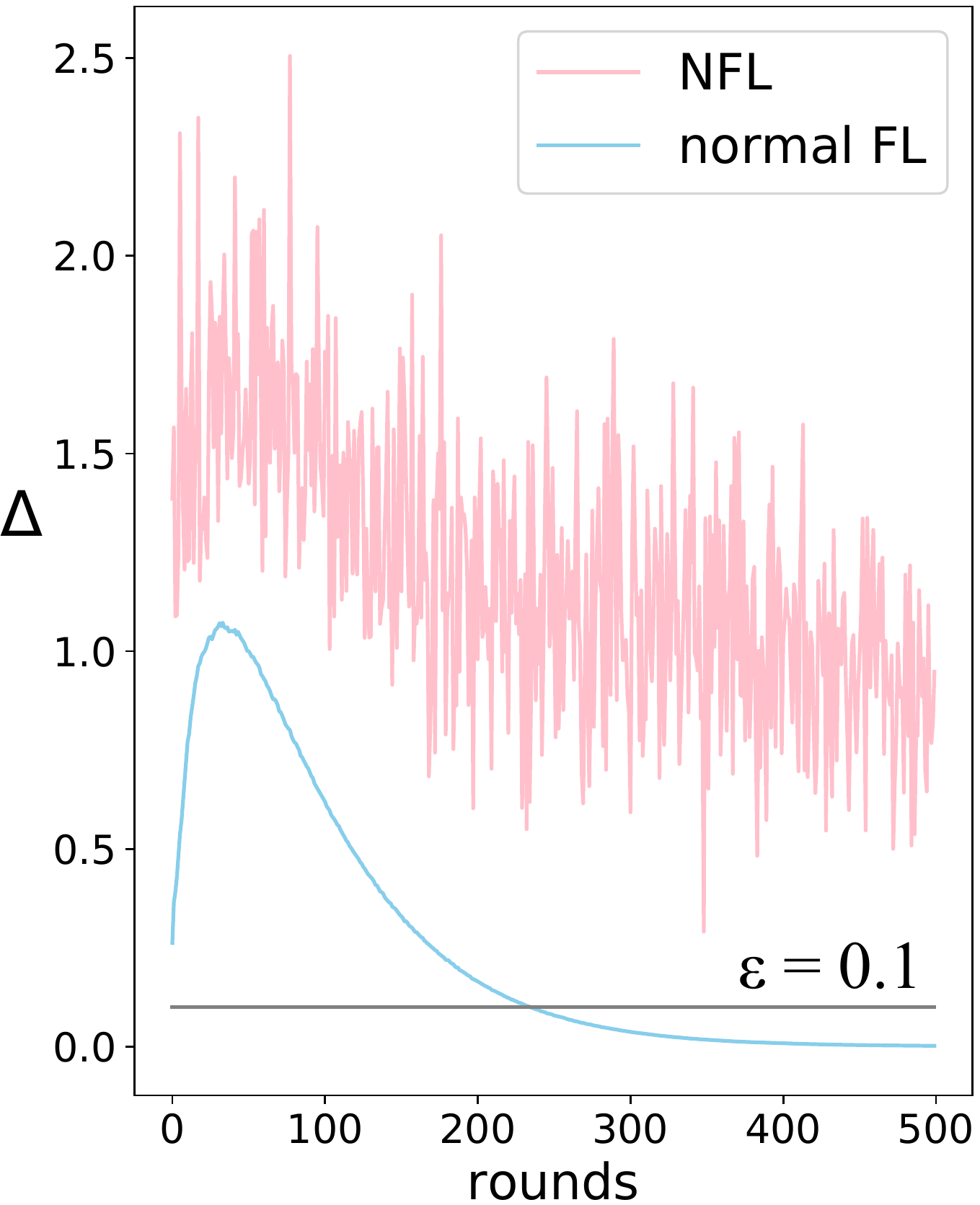}
         \caption{Results of $\Delta$.}
         \label{subfig:w_div_compare}
     \end{subfigure}
     \hfil
     \begin{subfigure}[b]{0.49\columnwidth}
         \includegraphics[width=\columnwidth,height=0.76\columnwidth]{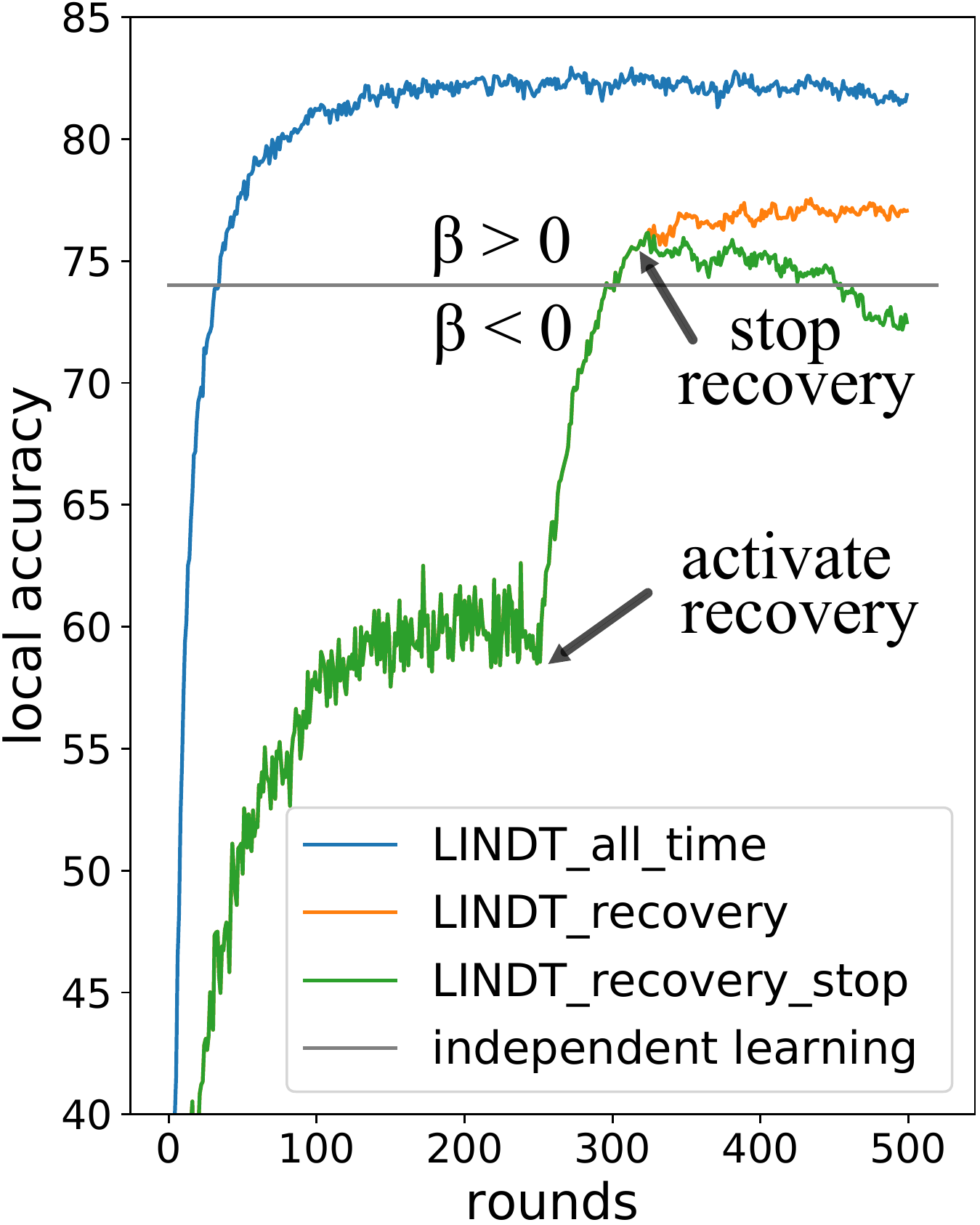}
         \caption{Run-time results of recovery.} 
         \label{subfig:recovery_stop}
     \end{subfigure}
}
\caption{NFL detection and recovery.} 
\label{fig:detect_recovery}
\end{figure}


To study NFL detection and recovery, we present the concrete results of $\Delta$ and local accuracy of each iterative round.
Fig.~\ref{fig:detect_recovery}(a) plots the results of $\Delta$ on CIFAR. Compared to normal FL process, the $\Delta$ value of learning in NFL (pink curve) fluctuates wildly above $\epsilon$, implying parameter disagreement on a global model among the clients. The results confirm the usefulness of $\Delta$ as a metric for NFL detection.

The run-time recovery process begins with FedAvg, and then sees NFL detected, and subsequently LINDT recovery activated. For reference, we also present the results of LINDT with all-time dual-model training.
To fully illustrate the power of LINDT, we set the recovery process to follow \textit{two stopping strategies} in multiple runs, where, (1) under the \textit{first strategy}, the system stops dual-model training (and degrades to FedAvg) after all clients participate in at least one round of dual-model training; whereas (2) in the \textit{second strategy}, the dual-model training persists once it is activated.

Fig.~\ref{fig:detect_recovery}(b) shows the results when $\epsilon=0.1$ and $r'=250$. It can be seen that when $r$ reaches 250, the system detects NFL, and then activates recovery immediately. As a result, the local accuracy increases rapidly, overtaking the grey line (indicating that the performance gain is turning positive). 

Interestingly, the accuracy path splits for different stopping strategies. When the system decides to stop recovery under the first strategy (green curve), we can see considerable drop in accuracy after the stopping point and in the end the system returns to NFL (as the green curve moves again below the grey line). In contrast, the orange curve keeps high above because the dual-model training never stops after activation. The all-time LINDT is the apparent winner. This, however, comes at the cost of 250 more rounds of dual-model training on all clients!
These results confirm the usefulness of dual-model training 
in recovery, namely dynamically improving local performance when NFL occurs.

\section{Conclusion}
This paper addressed the problem of negative federated learning (NFL).
We observed that significant divergence between the global optimum and local optimums in federated learning was an essential cause of NFL. Based on this observation, we introduced weight divergence along with a metric $\Delta$ for detecting NFL. Moreover, we proposed LINDT, a flexible local adaptation framework, 
for NFL detection and recovery. Experiments showed that LINDT achieved higher accuracy compared to previous solutions, and effectively handled various scenarios of NFL. We believe LINDT can be further improved by integrating other techniques in the relevant domain. This can be left for future work.

\bibliographystyle{plainnat} 
\bibliography{template}

\clearpage
\newpage

\appendix

\setcounter{secnumdepth}{2}


\section{Proof of Proposition 1} \label{A1}
According to Eq.~\ref{dp} and Eq.~\ref{wdivr}, we can get:
\begin{equation}\label{13}
\begin{aligned}
    & w\_div^{r} \\
    &= \frac{1}{K} \sum_{i \in C_r}
    \Vert \boldsymbol{w}_{i}^{r} - (\boldsymbol{w}^{r-1} + \frac{1}{K} \sum_{j \in C_r} Clip(\boldsymbol{w}_{j}^{r} - \boldsymbol{w}^{r-1} , S)) \\
    & \qquad \qquad \quad - \mathcal{N}(\boldsymbol{0}, \sigma^2 \boldsymbol{I})\Vert \\
    & \overset{1}{\leq} \frac{1}{K} \sum_{i \in C_r}
    \Vert \boldsymbol{w}_{i}^{r} - (\boldsymbol{w}^{r-1} + \frac{1}{K} \sum_{j \in C_r} Clip(\boldsymbol{w}_{j}^{r} - \boldsymbol{w}^{r-1} , S))  \Vert  \\
    & \quad + \frac{1}{K} \sum_{i \in C_r}  \Vert \mathcal{N}(\boldsymbol{0}, \sigma^2 \boldsymbol{I})\Vert \\
    & \overset{2}{\leq} \frac{1}{K} \sum_{i \in C_r}
    \Vert \boldsymbol{w}_{i}^{r} -  \frac{1}{K} \sum_{j \in C_r} \boldsymbol{w}_{j}^{r}  \Vert  +  \Vert  \mathcal{N}(\boldsymbol{0}, \sigma^2 \boldsymbol{I}))\Vert. \\
\end{aligned}
\end{equation}

Inequality 1 holds based on triangle inequality. Inequality 2 holds because the norm of the difference between two vectors is always larger than (or equal to) the norm after clipping one of the two vectors. In the following parts, we continue to bound  $\frac{1}{K} \sum_{i \in C_r} \Vert \boldsymbol{w}_{i}^{r} -  \frac{1}{K} \sum_{j \in C_r} \boldsymbol{w}_{j}^{r}  \Vert $: 
\begin{equation} \label{14}
\begin{aligned}
    &\frac{1}{K} \sum_{i \in C_r}
    \Vert \boldsymbol{w}_{i}^{r} -  \frac{1}{K} \sum_{j \in C_r} \boldsymbol{w}_{j}^{r}  \Vert  \\
    &= \frac{1}{K} \sum_{i \in C_r}  \Vert \frac{1}{K} \sum_{j \in C_r }  (\boldsymbol{w}_{i}^{r} - \boldsymbol{w}_{j}^{r})  \Vert \\
     & \leq \frac{1}{K^2} \sum_{i \in C_r} \sum_{j \in C_r \atop j \neq i}  \Vert   \boldsymbol{w}_{i}^{r} -   \boldsymbol{w}_{j}^{r}  \Vert. \\
\end{aligned}
\end{equation}

We rewrite $\Vert \boldsymbol{w}_{i}^{r} - \boldsymbol{w}_{j}^{r}  \Vert$ as $\Vert \boldsymbol{w}_{i}^{(r, T)} - \boldsymbol{w}_{j}^{(r, T)} \Vert$, denoting the value after $T$ steps of SGDs in the round $r$. Then, according to line c\ref{step12} in Algorithm \ref{fedavg_algorithm}, we have
\begin{equation}
\begin{aligned}
    &  \Vert  \boldsymbol{w}_{i}^{(r, T)} - \boldsymbol{w}_{j}^{(r, T)} \Vert  \\
    &=  \Vert \boldsymbol{w}_{i}^{(r, T-1)} - \eta \nabla_{\boldsymbol{w}} \mathbb{E}_{(x,y) \sim p_{i}} \ell (\boldsymbol{w}_{i}^{(r, T-1)}, x, y ) \\
    & \quad - (\boldsymbol{w}_{j}^{(r, T-1)} - \eta \nabla_{\boldsymbol{w}} \mathbb{E}_{(x,y) \sim p_{j}} \ell (\boldsymbol{w}_{j}^{(r, T-1)}, x, y ))  \Vert  \\
    &\overset{3}{\leq}  \Vert \boldsymbol{w}_{i}^{(r, T-1)} - \boldsymbol{w}_{j}^{(r, T-1)}  \Vert  \\
    & \quad + \eta  \Vert  \nabla_{\boldsymbol{w}} \mathbb{E}_{(x,y) \sim p_{i}} \ell (\boldsymbol{w}_{i}^{(r, T-1)}, x, y ) \\
    & \qquad \quad - \nabla_{\boldsymbol{w}} \mathbb{E}_{(x,y) \sim p_{j}} \ell (\boldsymbol{w}_{j}^{(r, T-1)}, x, y ) \Vert  \\
    &\overset{4}{\leq}  \Vert \boldsymbol{w}_{i}^{(r, T-1)} - \boldsymbol{w}_{j}^{(r, T-1)}  \Vert  \\
    & \quad + \eta  \Vert  \sum_{y \in Y} p_i(y) (\nabla_{\boldsymbol{w}} \mathbb{E}_{x|y} \ell (\boldsymbol{w}_{i}^{(r, T-1)}, x, y ) \\
    & \qquad \qquad \qquad \qquad - \nabla_{\boldsymbol{w}} \mathbb{E}_{x|y} \ell (\boldsymbol{w}_{j}^{(r, T-1)}, x, y ) ) \Vert \\
    & \quad + \eta  \Vert  \sum_{y \in Y} (p_i(y)-  p_j(y)) \nabla_{\boldsymbol{w}} \mathbb{E}_{x|y} \ell (\boldsymbol{w}_{j}^{(r, T-1)}, x, y) \Vert. \\
\end{aligned} \label{eq3}
\end{equation}

Inequality 3 holds due to triangle inequality. Inequality 4 holds because
\begin{equation}
\begin{aligned}
    &  \Vert  \nabla_{\boldsymbol{w}} \mathbb{E}_{(x,y) \sim p_{i}} \ell (\boldsymbol{w}_{i}^{(r, T-1)}, x, y)   \\
    & \quad - \nabla_{\boldsymbol{w}} \mathbb{E}_{(x,y) \sim p_{j}} \ell (\boldsymbol{w}_{j}^{(r, T-1)}, x, y ) \Vert  \\
    &=  \Vert \sum_{y \in Y} p_i(y) \nabla_{\boldsymbol{w}} \mathbb{E}_{x|y} \ell (\boldsymbol{w}_{i}^{(r, T-1)}, x, y ) \\
    & \qquad - \sum_{y \in Y} p_j(y) \nabla_{\boldsymbol{w}} \mathbb{E}_{x|y} \ell (\boldsymbol{w}_{j}^{(r, T-1)}, x, y ) \Vert  \\
    & =  \Vert  \sum_{y \in Y} p_i(y) (\nabla_{\boldsymbol{w}} \mathbb{E}_{x|y} \ell (\boldsymbol{w}_{i}^{(r, T-1)}, x, y ) \\
    & \qquad  \qquad  \qquad \quad  - \nabla_{\boldsymbol{w}} 
     \mathbb{E}_{x|y} \ell (\boldsymbol{w}_{j}^{(r, T-1)}, x, y ) )  \\
    & \quad  +  \sum_{y \in Y} (p_i(y)-  p_j(y)) \nabla_{\boldsymbol{w}} \mathbb{E}_{x|y} \ell (\boldsymbol{w}_{j}^{(r, T-1)}, x, y) \Vert \\
    & \leq  \Vert  \sum_{y \in Y} p_i(y) (\nabla_{\boldsymbol{w}} \mathbb{E}_{x|y} \ell (\boldsymbol{w}_{i}^{(r, T-1)}, x, y ) \\
    & \qquad  \qquad  \qquad \quad  - \nabla_{\boldsymbol{w}} 
     \mathbb{E}_{x|y} \ell (\boldsymbol{w}_{j}^{(r, T-1)}, x, y ) )  \Vert  \\
    & \quad  +  \Vert  \sum_{y \in Y} (p_i(y)-  p_j(y)) \nabla_{\boldsymbol{w}} \mathbb{E}_{x|y} \ell (\boldsymbol{w}_{j}^{(r, T-1)}, x, y) \Vert. \\
\end{aligned}
\end{equation}
   
We let $g_{max}(\boldsymbol{w}) = \max_{y \in Y} \Vert \nabla_{\boldsymbol{w}} \mathbb{E}_{x|y} \ell (\boldsymbol{w}, x, y ) \Vert $ 
and assume $ \nabla_{\boldsymbol{w}} \mathbb{E}_{x|y} \ell (\boldsymbol{w}, x, y )$ is $\lambda_{x|y}$-Lipschitz for every label $y \in Y$. We can simplify  Eq.~\ref{eq3} by 
\begin{equation}
\begin{aligned}
    &  \Vert  \boldsymbol{w}_{i}^{(r, T)} -  \boldsymbol{w}_{j}^{(r, T)}  \Vert  \\
    & \leq (1+\eta \sum_{y \in Y} \lambda_{x|y} p_i(y) )  \Vert \boldsymbol{w}_{i}^{(r, T-1)} - \boldsymbol{w}_{j}^{(r, T-1)}  \Vert  \\
    & \quad + \eta g_{max}(\boldsymbol{w}_{j}^{(r, T-1)}) \sum_{y \in Y}  \Vert p_i(y)-  p_j(y) \Vert.
\end{aligned} \label{eq4}
\end{equation}

Based on Eq.~\ref{eq4}, let $a_i = 1+\eta \sum_{y \in Y} \lambda_{x|y} p_i(y)$, by induction, we have 

\begin{equation}
\begin{aligned}
    & \Vert  \boldsymbol{w}_{i}^{(r, T)} -  \boldsymbol{w}_{j}^{(r, T)}  \Vert  \\
    &\leq a_i  \Vert \boldsymbol{w}_{i}^{(r, T-1)} - \boldsymbol{w}_{j}^{(r, T-1)}  \Vert  \\
    & \quad  + \eta g_{max}(\boldsymbol{w}_{j}^{(r, T-1)}) \sum_{y \in Y}  \Vert p_i(y)-  p_j(y) \Vert  \\
    & \leq (a_i)^2  \Vert \boldsymbol{w}_{i}^{(r, T-2)} - \boldsymbol{w}_{j}^{(r, T-2)}  \Vert  \\
    & \quad  + \eta (g_{max}(\boldsymbol{w}_{j}^{(r, T-1)}) \sum_{y \in Y}  \Vert p_i(y)-  p_j(y) \Vert  \\
    & \quad  \qquad  + a_i g_{max}(\boldsymbol{w}_{j}^{(r, T-2)}) \sum_{y \in Y}  \Vert p_i(y)-  p_j(y) \Vert )   \\
    &\leq (a_i)^{T}  \Vert \boldsymbol{w}_{i}^{(r, 0)} - \boldsymbol{w}_{j}^{(r, 0)}  \Vert  \\
    & \quad + \eta \sum_{y \in Y}  \Vert p_i(y)-  p_j(y) \Vert  \sum_{t=0}^{T-1} (a_i)^{t} g_{max}(\boldsymbol{w}_{j}^{(r, T-1 -t)}).
\end{aligned}
\end{equation}

Since $\boldsymbol{w}_{i}^{(r, 0)} = \boldsymbol{w}_{j}^{(r, 0)} = \boldsymbol{w}^{r-1}$, we have
\begin{equation} \label{18}
\begin{aligned}
    & \Vert  \boldsymbol{w}_{i}^{(r, T)} -  \boldsymbol{w}_{j}^{(r, T)}  \Vert  \\
    &\leq  \eta \sum_{y \in Y}  \Vert p_i(y)-  p_j(y) \Vert  \sum_{t=0}^{T-1} (a_i)^{t} g_{max}(\boldsymbol{w}_{j}^{(r, T-1 -t)}).
\end{aligned}
\end{equation}

According to Eq.~\ref{13}-\ref{14} and Eq.~\ref{18}, we obtain 
\begin{equation}\label{19}
\begin{aligned}
     w\_div^{r} & \leq  \Vert  \mathcal{N}(\boldsymbol{0}, \sigma^2 \boldsymbol{I}))\Vert 
    + \frac{\eta}{K^2} \sum_{i \in C_r} \sum_{j \in C_r \atop j \neq i}\\
    & \quad  \sum_{y \in Y}  \Vert p_i(y)-  p_j(y) \Vert  \sum_{t=0}^{T-1} (a_i)^{t} g_{max}(\boldsymbol{w}_{j}^{(r,T-1 -t)}).
\end{aligned}
\end{equation}

Proof ends. 

\section{Details of LINDT} \label{A2} 
\subsection{Attaching Module}
\begin{figure}[h]
\centering
\includegraphics[width=0.6\columnwidth]{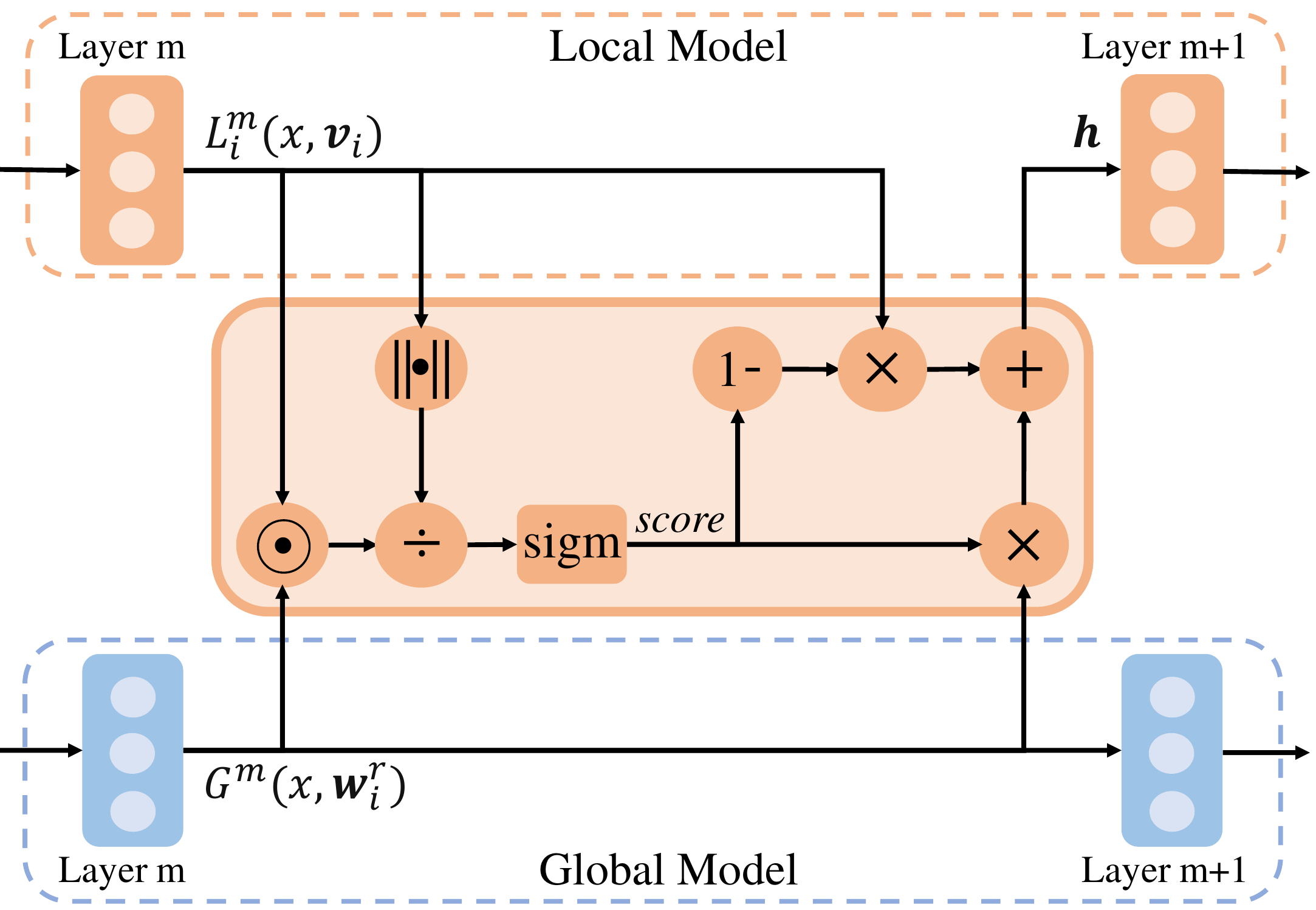}
\caption{The detailed structure of an attaching module at layer $m$, where $m=1,2,\ldots,M-1$.} 
\label{attaching_module}
\end{figure}

Fig.~\ref{attaching_module} shows the internal structure of the attaching module used for layer-wise intertwinement between the global model and the local model. Given a total of $M$ layers in the dual-model, the attaching module at the $m$-th layer is fed with two hidden representations of a data sample $x$, denoted by $G^{m}(x, \boldsymbol{w}_{i}^{r})$ and $L_{i}^{m}(x, \boldsymbol{v}_i)$, where the former is the output from the $m$-th layer of the global model, and the latter is the output from the $m$-th layer of the local model on client $i$. The module makes the following computations: It first makes a dot product between these two representations, which is represented by $\bigodot$. Then, it normalizes the dot product by the norm of $L_{i}^{m}(x, \boldsymbol{v}_i)$. Next, the module calculates the $score$ as defined in Eq.~\ref{attention1} of Section \ref{sec:intertwinement} via a $sigmoid$ function (represented by $
sigm$). Finally, as defined in Eq.~\ref{attention2}, the attaching module adds the product of $G^{m}(x, \boldsymbol{w}_{i}^{r})$ and $score$ with the product of $L_{i}^{m}(x, \boldsymbol{v}_i)$ and $1-score$, generating the intertwined representation $\boldsymbol{h}$ to be fed into the ($m+1$)-th layer of the local model.

\subsection{The LINDT Algorithm}

The pseudo code of LINDT (NFL detection and recovery mode) is shown in Algorithm \ref{alg:Framwork}. The other running mode (all-time LINDT) can be obtained by making trivial amendments to the algorithm, i.e., initializing $flag=True$ at line s1.

\renewcommand{\algorithmicrequire}{ \textbf{Server executes:}} 
\renewcommand{\algorithmicensure}{ \textbf{Clients execute:}} 

\begin{algorithm}[!htb] 
\small
\caption{LINDT: Layer-wise INterwined Dual Training}
\label{alg:Framwork} 

\textbf{Input:} 
A set of clients $i \in \{1,\ldots,N\}$. The number of clients that perform computation in each round, $K$. Local mini-batch size, $B$. The number of local epochs, $E$. Learning rate, $\eta$. 
Upper bound to the norm of weight updates, $S$. Std for generating Gaussian noise, $\sigma$. A threshold for $\Delta$, $\epsilon$. A threshold restricting the rounds in which $\Delta > \epsilon$,  $r'$.  The number of rounds in which $\Delta > \epsilon$, $count$. A flag denoting whether NFL is detected, $flag$.

\algrenewcommand{\alglinenumber}[1]{\footnotesize {s#1}:}%

\begin{algorithmic}[1] 
\Require 
\label{step1}
\State initialize $\boldsymbol{w}^{0}$, $count=0$, $flag = False$
\For{each round $r$=1,2,\ldots}
    \State  broadcast the latest weight vectors $\boldsymbol{w}^{r-1}$ to clients
    \State wait until receiving $K$ locally-updated weight vectors $\boldsymbol{w}^{r}_{i}$ 
    \Statex \hspace{\algorithmicindent}from a set of active clients $C_r$
    \State generate $noise = \mathcal{N}(\boldsymbol{0}, \sigma^2 \boldsymbol{I})$ and aggregate local updates
    \Statex \hspace{\algorithmicindent}under differential privacy protection: 
    \Statex \hspace{\algorithmicindent} $ \boldsymbol{w}^{r} \leftarrow  \boldsymbol{w}^{r-1} +  \frac{1}{K} \sum_{i \in C_r} Clip(\boldsymbol{w}_{i}^{r} - \boldsymbol{w}^{r-1} , S) +  noise$ 
    \State calculate $w\_div^{r} = \frac{1}{K} \sum_{i \in C_r}\Vert\boldsymbol{w}_{i}^{r} - \boldsymbol{w}^{r}\Vert$
    \State calculate $\Delta  = w\_div^{r} - \Vert noise \Vert $
    \If{$\Delta > \epsilon$} 
        \State $count = count+1$
    \EndIf
    \If{$count > r'$ and $flag == False$}
        \State report NFL is detected and set $flag = True$
    \EndIf
\EndFor
\end{algorithmic}

\algrenewcommand{\alglinenumber}[1]{\footnotesize {c}#1:}%
\begin{algorithmic}[1] 
\Ensure 
\For{each round $r$=1,2,\ldots}
    \State download the latest weight vectors  $\boldsymbol{w}^{r-1}$ from server
    \State $\boldsymbol{w}_{i}^{r} \leftarrow \boldsymbol{w}^{r-1}$;
    \If{$flag == True$ }
        \If{$\boldsymbol{v}_{i}$ has not been initialized}
            \State  initialize $\boldsymbol{v}_{i}$ 
        \EndIf
        \State \textbf{layer-wise intertwine} the received global model 
        \Statex \hspace{\algorithmicindent} \hspace{\algorithmicindent} $G(\cdot, \boldsymbol{w}_{i}^{r})$ with local model $L_i(\cdot, \boldsymbol{v}_{i})$
    \EndIf
    \State $\mathcal{B}_i \leftarrow$ split local data $D_i$ into batches of size $B$
    \For{each local epoch from 1 to $E$}
        \For{batch  $b_i \in \mathcal{B}_i$}
            
            \If{$flag == True$}
                \State  $\boldsymbol{w}_{i}^{r} \leftarrow \boldsymbol{w}_{i}^{r} - \eta \nabla_{\boldsymbol{w}_{i}^{r}} \ell (\boldsymbol{w}_{i}^{r}, \boldsymbol{v}_{i})$
                \State  $\boldsymbol{v}_{i} \leftarrow \boldsymbol{v}_{i} - \eta \nabla_{\boldsymbol{v}_{i}} \ell (\boldsymbol{w}_{i}^{r}, \boldsymbol{v}_{i})$
            \Else   
                \State  $\boldsymbol{w}_{i}^{r} \leftarrow \boldsymbol{w}_{i}^{r} - \eta \nabla_{\boldsymbol{w}_{i}^{r}} \ell (\boldsymbol{w}_{i}^{r}, \boldsymbol{v}_{i})$
            \EndIf
        \EndFor
    \EndFor
    \State return $\boldsymbol{w}_{i}^{r}$ to server
\EndFor
\end{algorithmic}
\end{algorithm}

\section{Experiment Details and Extra Results}\label{experiment_appedix}
Here we provide all the details regarding experimental setup, dataset preprocessing, model architectures, model training, and
extra experimental results. Our anonymized code, which is implemented by Pytorch 1.2.0 \citep{paszke2017automatic}, is attached in the supplementary material. All experiments are conducted on a single machine with 2 GeForce GTX 1080 Ti GPUs.

\subsection{Federated Environment Setup} \label{C1}

\subsubsection{Dataset and client data allocation} We use two benchmark datasets in our experiment, CIFAR-10 \citep{Krizhevsky2009LearningML} and Shakespeare \citep{Caldas2018LEAFAB}. 
For CIFAR-10, we test the following five schemes to allocate its 50,000 training data and 10,000 testing data to $N = 100$ clients:
\begin{itemize}
    \item IID: the complete dataset is shuffled and then each client is randomly allocated with the \textit{same} amount of data over 10 classes. 
    
    \item non-IID(\textit{10}): the complete dataset is sorted by labels and then each client is randomly allocated with a \textit{different} amount of data from \textit{10} classes.
    
    \item non-IID(\textit{5}): the complete dataset is sorted by labels and each client is randomly allocated with a \textit{different} amount of data from only \textit{5} classes.
    
    \item non-IID(\textit{2}): the complete dataset is sorted by labels and each client is randomly allocated with a \textit{different} amount of data from only \textit{2} classes.
    
    \item non-IID(\textit{Mixed}): the complete dataset is sorted by labels and clients are divided into three groups of size 50, 30, 20. Then we set up a case, where 50 clients have examples from 10 classes, 30 clients have examples from 5 classes, and the remaining 20 clients have examples from 2 classes.
\end{itemize}
In all above schemes except for IID, the amount of data on each client is set to follow a log-normal distribution. Among these schemes, non-IID(\textit{Mixed}) is the default one for allocating CIFAR-10. 

%
For Shakespeare, we allocate each speaking role to one client \citep{Caldas2018LEAFAB}. Following \citet{Wang2020Federated}, we preprocess this dataset by filtering out the clients with less than 10,000 datapoints and sampling a random subset of $N = 66$ clients. We allocate 90\% of the data for training and the remaining for testing. The sampled dataset contains in total 1,053,880 training data and 117,134 testing data. We term this allocation scheme as \textit{`by role'} and set it as the default for allocating Shakespeare dataset. 

We also form a \textit{balanced} version of data allocation on Shakespeare. The complete dataset is shuffled and then each of $66$ clients is randomly allocated with the \textit{same} amount of data over 80 classes. Table \ref{dataStatistic} summarizes the statistical information of client data under different allocation schemes.

\begin{table}[h]
  \small
  \centering
  \begin{tabular}{cccc}
    \toprule
    \multirow{2}[4]{*}{Dataset} & \multirow{2}[4]{*}{Allocation} & \multicolumn{2}{c}{\# Data per client} \\
\cmidrule{3-4}      &       &    mean  & std \\
    \midrule
    \multirow{5}[1]{*}{CIFAR-10} & IID & \multirow{5}[1]{*}{600}  & 0 \\
          & non-IID(10) &   & 572 \\
          & non-IID(5) &  & 811 \\
          & non-IID(2) &    & 855 \\
          & non-IID(\textit{Mixed}) *  &  & 732 \\
   \midrule       
   \multirow{2}[1]{*}{ Shakespeare} & by role * & 17742 & 8997 \\
   & balanced & 17741 & 0 \\
    
    \bottomrule
    \end{tabular}%
    \caption{Statistical information of client data under different data allocation schemes. In each scheme, every client has the same distributions on its own train and test data. *: the default allocation scheme in main experiments.}
  \label{dataStatistic}%
\end{table}%

\subsubsection{Simulation of client inactivity} 
We simulate client inactivity by random sampling only $10\%$ of all clients ($K/N=10\%$) in every round to participate in federated model training. Among the selected clients in every round, there exist several malicious attackers poisoning the FL-trained global model. The details of attacks are given in the following text.

\subsubsection{Simulation of attacks}
In our experiments, we randomly select some of the clients to be malicious attackers who poison the FL-trained model via reporting the model weights updated on a backdoor dataset \citep{Bagdasaryan2020HowTB,Bhagoji2018}.
\begin{itemize}
    \item For CIFAR-10 experiments, we set 20\% of clients to be attackers (20 attackers out of the total $N=100$ clients) and randomly sample $20\% * K/N * N$ attackers to participate in every round (i.e, 2 attackers sampled per round to work with the other 8 normal clients). The simulated attackers falsely label the images of deer and dog as horse and frog respectively in their backdoor dataset. 
    
    \item For Shakespeare experiments, the proportion of attacker is set as 16\% (11 attackers out of the total $N=66$ clients) and we randomly sample $16\% * K/N * N$ attackers to participate in every round (i.e, 1 attackers sampled per round to work with the other 5 normal clients). Attackers aim to manipulate the global model to predict the next character of all input sentences to be `\textbackslash n', which makes no sense in normal speech. 
\end{itemize}

When training the attacker’s model, we follow \citet{Gu2017BadNetsIV} and \citet{Bagdasaryan2020HowTB} to mix backdoor samples with normal samples (labeled by ground truth) in every training batch (3 backdoor samples per batch of size 10). Although such operation weakens the attacks, it prevents the malicious updates from being easily identified by the central server. Note that in the Shakespeare experiments, such data mixing scheme does not make sense as the attacker simply produces `\textbackslash n' in a row with no possession of normal samples. Therefore, \textit{we only apply this operation in CIFAR-10 experiments.} Same as \citet{Bagdasaryan2020HowTB}, we also allow attackers to run more epochs of local training per round, which benefits the manipulation of the global model to be more overfitting their backdoor data. For all experiments, attackers run $5$ local epochs per round (vs. $1$ local epoch for the normal clients).

\subsubsection{Simulation of differential privacy protection}
We simulate the differential privacy protection on the server following the scheme in \citep{McMahan2018LearningDP}. We set the hyperparameters in Eq.~\ref{dp} as $S=15$, $\sigma= 0.001$.

\begin{table}[t]
  \small
  \centering
  \begin{tabular}{ccc}
    \toprule
      & CNN & LSTM \\
    \midrule
    \# Parameters & 940,362 & 819,920\\
    Optimizer & SGD & SGD\\
    learning rate ($\eta$) & 0.1 & 1.47\\
    Learning rate decay & 0.992 & 0.992\\
    dropout rate & 0.5 & 0.2 \\
    max norm of gradient & 5 & 5 \\
    $\#$rounds & 500 & 150\\
    train batch size ($B$) & 10 & 50\\
    test batch size & 128 & 128\\
    local epochs ($E$) & 1 & 1 \\
    \bottomrule
    \end{tabular}%
    \caption{Hyper-parameters used in training the two models.}
  \label{hyper}%
\end{table}%
\subsection{Models and Hyperparameters} \label{C2}
Our experiments train the following  popular neural models on the two datasets. 
\begin{itemize}
    \item On CIFAR-10, we train a CNN which is composed of two 5x5 convolution layers (the first with 32 channels, the second with 64, each followed with 2x2 max pooling), two fully connected layers with ReLu activation and respectively 512 units and 128 units, as well as a final softmax output layer (a total of 940,362 parameters).
    
    \item On Shakespeare dataset,  we train a stacked character-level LSTM language model, which, after reading a sequence of characters in a line, predicts the next character \citep{McMahan2017CommunicationEfficientLO}. The model takes a series of characters as input and embeds each of these into a learned 8 dimensional space. The embedded characters are then processed through 2 LSTM layers, each with 256 nodes. Finally the output of the second LSTM layer is sent to a softmax output layer with one node per character. The full model has 819,920 parameters, and we set the unroll length of every input as 80 characters.
\end{itemize} 
Table \ref{hyper} summarizes the hyper-parameters used in training the models.

\subsection{Hyperparameters in Previous FL Methods} \label{C3}
In section \ref{compare_previous} we compared our method with FedAvg \citep{McMahan2017CommunicationEfficientLO}, FedProx \citep{Sahu2018FederatedOF}, TrimmedMean \citep{pmlr-v80-yin18a}, FB \citep{Wang2019FederatedEO}, and APFL \citep{Deng2020AdaptivePF}. Some of these methods use additional hyperparameters as listed in the following:
\begin{itemize}
    \item \textit{FedProx}: the coefficient $\mu$ associated with the proxy term. For all experiments, $\mu=0.001$.
    \item \textit{FB}: the learning rate $lr$ and the epochs $E'$ for fine-tuning model. For experiments on CIFAR-10, $lr = 0.001$ and $E' = 200$, while for experiments on Shakespeare, $lr = 0.01$ and $E' = 40$.
    \item \textit{APFL}: the coefficient for mixing global and local model on each client. For all experiments, such coefficient is initialized by 0.01 and then adaptive updated during training as mentioned in  \citep{Deng2020AdaptivePF}.
\end{itemize}


\subsection{Local Accuracy over Training Rounds} \label{C5}
We extend the results reported in Section \ref{compare_previous} by showing the local accuracy of different methods in each training round. We present the accuracy curve of FedAvg, APFL, and LINDT, while omitting others due to their similar behavior as FedAvg. 

\begin{figure}[t]
\centering
\begin{subfigure}[t]{0.49\columnwidth}
    \includegraphics[width=\columnwidth,height=0.75\columnwidth]{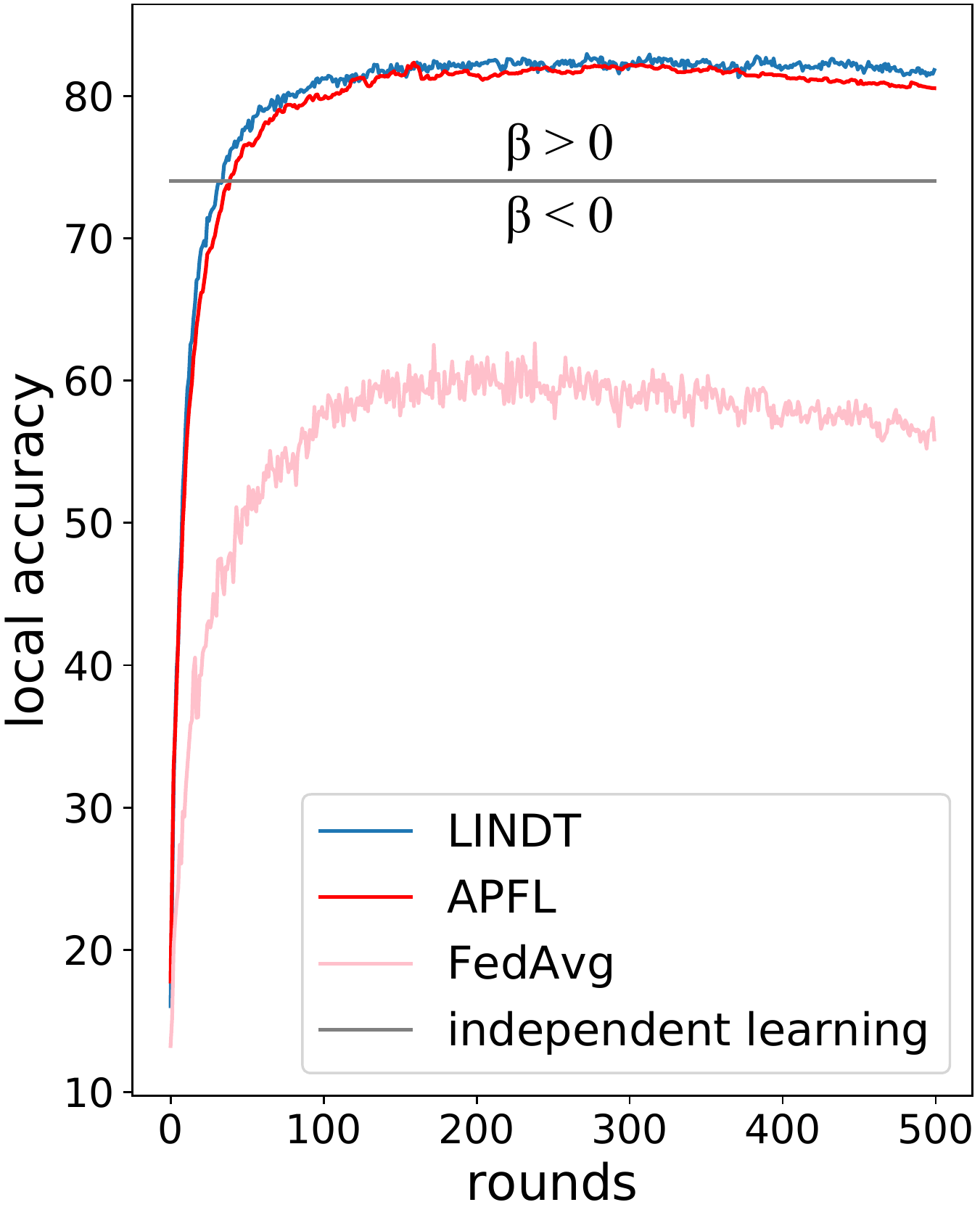}
    \caption{CIFAR-10}
\end{subfigure}
\hfil
\begin{subfigure}[t]{0.49\columnwidth}
    \includegraphics[width=\columnwidth,height=0.75\columnwidth]{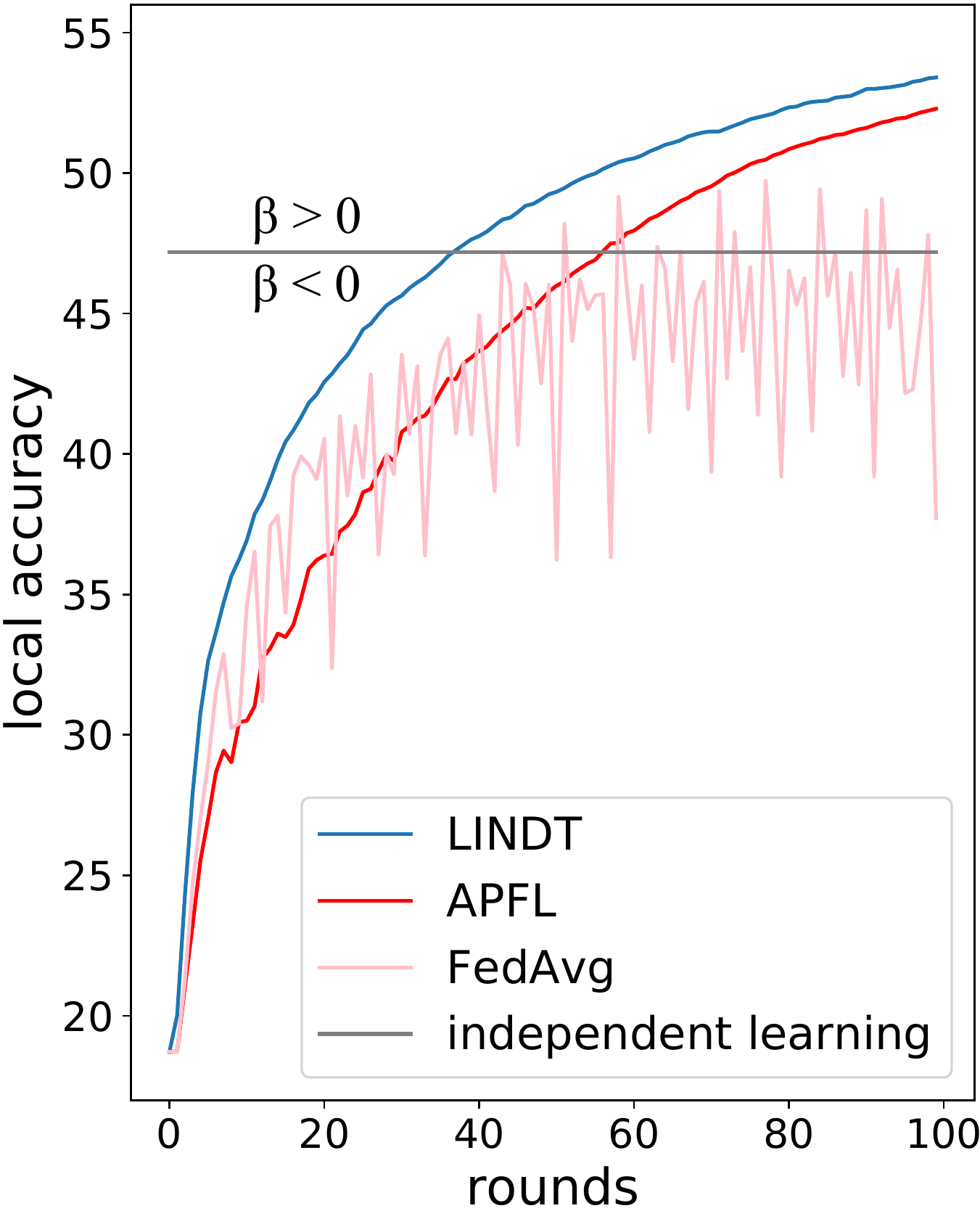}
    \caption{Shakespeare}
\end{subfigure}

\caption{Comparison of local accuracy over training rounds.
} 
\label{lacc}
\end{figure}

As presented in Fig.~\ref{lacc}, the local accuracy of FedAvg fluctuates more widely than that of the other two approaches. FedAvg only trains for all clients a single globally-shared model, which continuously suffers the negative effects from system attacks, different local data distributions, etc. As a result, such a global model can hardly reach a stable accuracy on individual clients. On the contrary, both LINDT and APFL learn an additional local model on every client while training the global model. The local model can fit the local data better and is exempted from the negative effects inherent in FL. Therefore, LINDT along with APFL can converge much quicker than FedAvg and result in a positive performance gain ($\beta>0$) for the clients participating in FL. 
LINDT further outperforms APFL by converging to a higher local accuracy, which is beneficial for encouraging more clients to contribute to FL.

\subsection{Extra Experimental Results on Shakespeare} \label{A12}

In this section, we present (1) the results of tuning environment parameters and (2) the results of $\Delta$ on the Shakespeare dataset. The same results for CIFAR-10 have been reported in Section \ref{sec:tuning} and Section \ref{sec:detect_recovery} respectively.

\subsubsection{Tuning environment settings}
The tuning results in Table \ref{noniid2}-\ref{sigma2} present the similar trends/patterns as discussed in Section \ref{sec:tuning}. We observe NFL is still prevalent in the entire parameter space that we tune. In cases of NFL, the global federated model can hardly reach high accuracy. However, the dual-model in LINDT can always ensure a positive gain in local accuracy ($\beta>0$), which is important for motivating clients to participate in FL.

Table \ref{noniid2} shows the negative impact on FL from the difference in client data distributions. A decrease in central and local accuracy appears in FedAvg when the data distributions become more different across clients. But compared to the results reported on CIFAR (Table \ref{noniid}), such decrease is not significant because the `by role' allocation scheme does not enlarge the difference in client data distributions very much (e.g. all clients have the same class labels under both `balanced' and `by role' schemes).
In contrast to FedAvg, LINDT achieves much higher local accuracy along with an improved central accuracy, which confirms its effectiveness to well-adapt to variable data distributions.

\begin{table}[!t]
  \centering
  \small %
    \begin{tabular}{l|cc|cc|cc}
    \toprule
    \makecell[c]{Data} & \multicolumn{2}{c}{Central ACC} \vline  & \multicolumn{2}{c}{Local ACC} \vline & \multicolumn{2}{c}{$\beta$}\\
    \makecell[c]{Alloc.} & FedAvg & LINDT  & FedAvg & LINDT & FedAvg & LINDT \\
    \midrule
    
    balanced & 44.77 & \textbf{52.57} & 45.32 & \textbf{52.50} & +2.03 & \textbf{+9.21} \\ 
    
    by role* & 43.60 & \textbf{49.17} & 44.26 & \textbf{52.84} & -2.93 & \textbf{+5.65} \\
    \bottomrule
    \end{tabular}%
    \caption{Varying data distributions among clients (Shakespeare).}
    
  \label{noniid2}%
\end{table}%

\begin{table}[t]
  \centering
  \small %
    \begin{tabular}{c|cc|cc|cc}
    \toprule
    \multirow{2}{*}{$K/N$} & \multicolumn{2}{c}{Central ACC} \vline & \multicolumn{2}{c}{Local ACC} \vline &  \multicolumn{2}{c}{$\beta$}  \\
     & FedAvg & LINDT  & FedAvg & LINDT  & FedAvg & LINDT \\
    \midrule
        
    90\%    & 45.69 & \textbf{49.89} & 45.72  & \textbf{52.93} & -1.47 & \textbf{+5.74} \\ 
         
    30\%    & 40.30 & \textbf{44.27} & 40.72 & \textbf{51.67} & -6.47 & \textbf{+4.48} \\ 
         
    10\%*  & 43.60 & \textbf{49.17} & 44.26 & \textbf{52.84} & -2.93 & \textbf{+5.65} \\
    \bottomrule
    \end{tabular}%
    \caption{Varying the ratio of active clients in each round (Shakespeare).
    }

  \label{active2}%
\end{table}%

\begin{table}[!t]
  \centering
  \small 
    \begin{tabular}{c|cc|cc|cc}
    \toprule
    \multirow{2}{*}{Attack} & \multicolumn{2}{c}{Central ACC} \vline & \multicolumn{2}{c}{Local ACC} \vline &  \multicolumn{2}{c}{$\beta$}  \\
     & FedAvg & LINDT  & FedAvg & LINDT  & FedAvg & LINDT \\
    \midrule
    0\%   & 52.93 & \textbf{53.67} & 53.42 & \textbf{54.07} & +6.16 & \textbf{+6.81} \\ 
    
    16\%*   & 43.60 & \textbf{49.17} & 44.26 & \textbf{52.84} & -2.93 & \textbf{+5.65} \\
    
    33\%   & 25.20 & \textbf{30.97} & 25.75 & \textbf{51.21} & -21.37 & \textbf{+4.09} \\ 
         
    \bottomrule
    \end{tabular}%
    \caption{Varying the proportion of attackers in each round (Shakespeare).}

  \label{attack2}%
\end{table}%

Table \ref{active2} and Table \ref{attack2} confirm that client inactivity and attacks do have negative impacts on FL. The conclusion on these results is similar to that in Section \ref{sec:tuning}. But note that in Shakespeare experiments, the change in central and local ACC is not monotonic to the change in the ratio of active clients in each rounds. This is probably due to the random disparities in reported parameters from more active clients. 

Table \ref{sigma2} presents the negative impact caused by the noises introduced by different privacy. A slight change occurs in both central and local accuracy as the noises increase quantitatively from $\sigma=0.001$ to $\sigma=0.01$. Compared to the results on CIFAR (Table \ref{sigma}), the change here is not considerable and not strictly monotonic. This indicates the better resilience of the federated LSTM against the noise from differential privacy than the federated CNN. 
In contrast to FedAvg, LINDT produces much better central accuracy and local accuracy. These results show that LINDT is more resilient against the large noise and can make good adaptation on client data.
\begin{table}[!t]
  \centering
  \small 
    \begin{tabular}{c|cc|cc|cc}
    \toprule
    \multirow{2}{*}{$\sigma$} & \multicolumn{2}{c}{Central ACC} \vline & \multicolumn{2}{c}{Local ACC} \vline &  \multicolumn{2}{c}{$\beta$}  \\
     & FedAvg & LINDT  & FedAvg & LINDT  & FedAvg & LINDT \\
    \midrule
   
    0.001*  & 43.60 & \textbf{49.17} & 44.26 & \textbf{52.84} & -2.93 & \textbf{+5.65} \\
    
    0.003   & 43.90 & \textbf{52.77} & 44.43 & \textbf{53.21} & -2.76 & \textbf{+6.02} \\ 
    
    0.005   & 43.70 & \textbf{49.07} & 44.21 & \textbf{52.66} & -2.98 & \textbf{+5.47} \\ 
    
    0.007   & 43.80 & \textbf{51.10} & 44.49 & \textbf{52.89} & -2.70 & \textbf{+5.70} \\ 
    
    0.01  & 43.10 & \textbf{47.69} & 43.52 & \textbf{52.43} & -3.67 & \textbf{+5.24} \\

    \bottomrule
    \end{tabular}%
    \caption{Varying the \textit{std} ($\sigma$) of noises for differential privacy (Shakespeare). }
  \label{sigma2}%
\end{table}%

\subsubsection{Results of $\Delta$}
\begin{figure}[t]
\centering
\begin{subfigure}[t]{0.49\columnwidth}
    \includegraphics[width=\columnwidth,height=0.75\columnwidth]{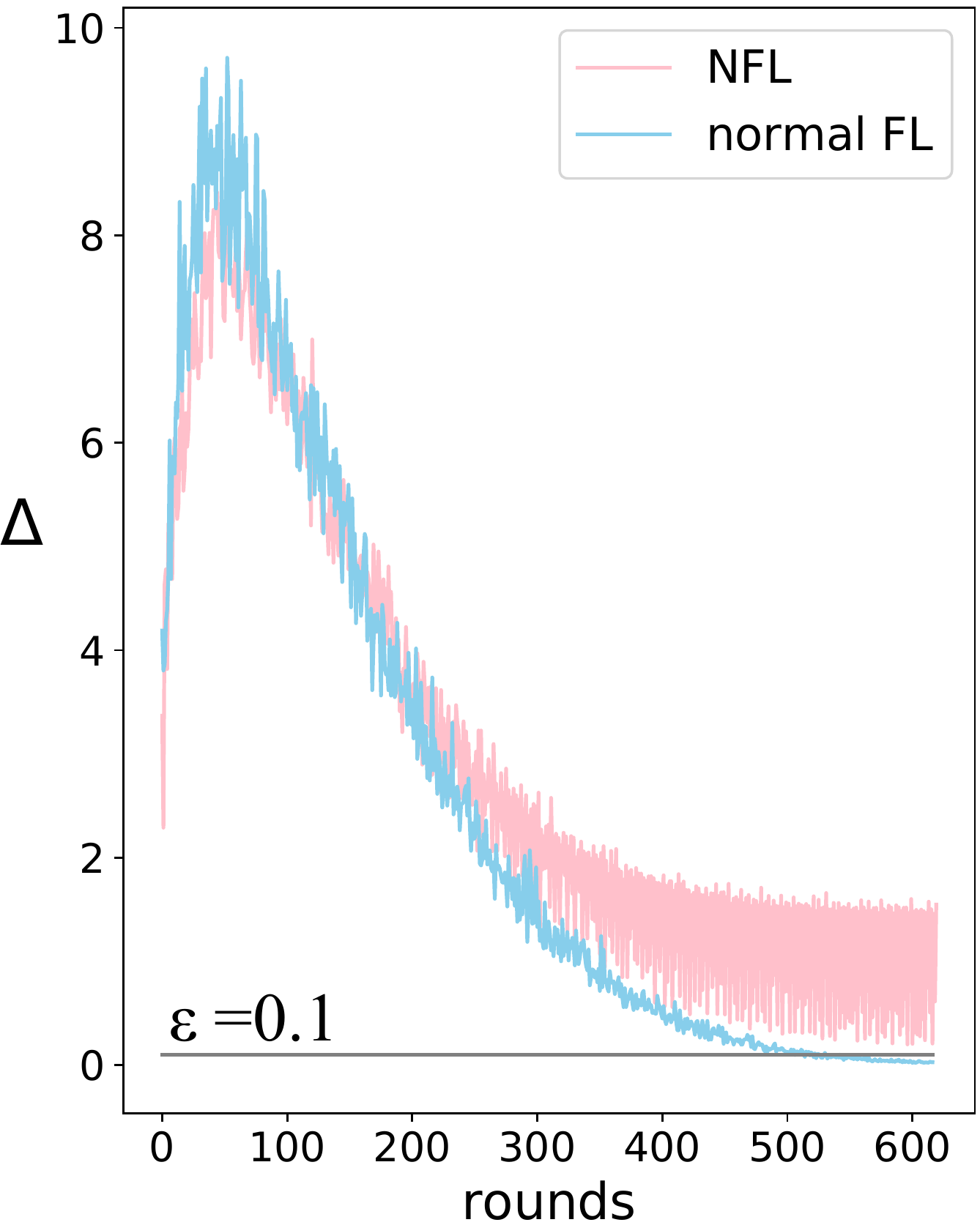}
    \caption{Over all training rounds.}
\end{subfigure}
\hfil
\begin{subfigure}[t]{0.49\columnwidth}
    \includegraphics[width=\columnwidth,height=0.75\columnwidth]{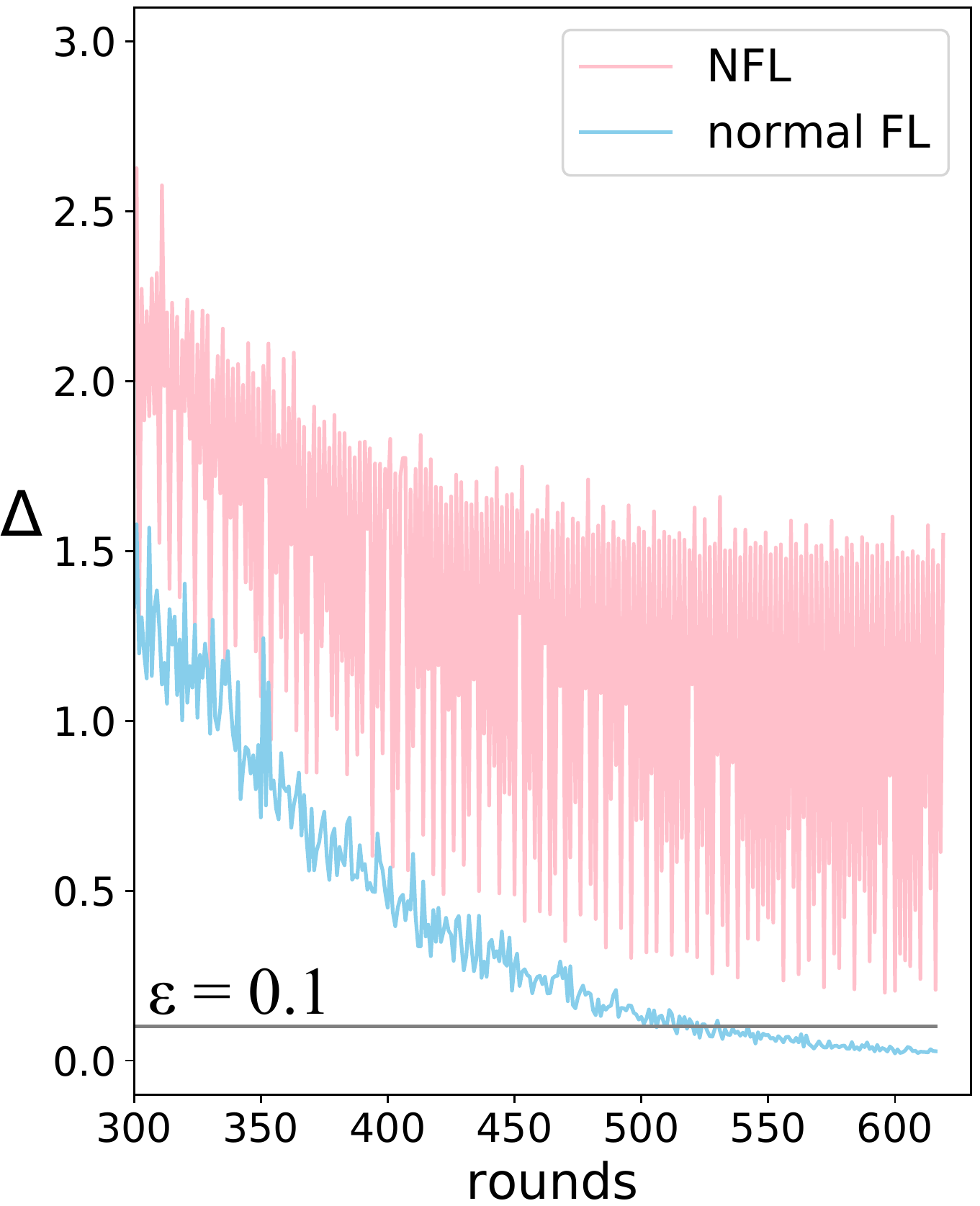}
    \caption{After 300 training rounds.} 
\end{subfigure}
\caption{Results of $\Delta$ on Shakespeare. The NFL process runs in the default NFL environment as specified in Table \ref{task_profile}. The normal FL process runs in an environment where no attack exists and the rest parameters are set as default.
} 
\label{fig:delta_on_shake}
\end{figure}

Fig.~\ref{fig:delta_on_shake} presents the results of $\Delta$ on Shakespeare. Similar as the results on CIFAR (Fig.~\ref{fig:detect_recovery}(a)), the $\Delta$ value in the NFL process still fluctuates widely above $\epsilon=0.1$, even if we take more than 500 rounds of training. In contrast, $\Delta$ in a normal FL process gradually approaches zero. These results further confirm the usefulness of $\Delta$ as a metric for NFL detection.

\subsection{What is $\Delta$ Not Intended For?}
In this section, we further clarify the intended usage of $\Delta$ by correcting a \textit{pitfall}.
Remember that in the main paper, we have already demonstrated the effective use of $\Delta$ for detecting NFL. When looking at the detection/recovery process shown in Fig.~\ref{fig:detect_recovery}(b), one may think that the value of $\Delta$ in NFL, after activation of recovery, may rapidly approach zero, just like it does in normal FL.
However, this is NOT true. 

\begin{figure}[t]
\centering
\begin{subfigure}[t]{0.49\columnwidth}
    \includegraphics[width=\columnwidth,height=0.75\columnwidth]{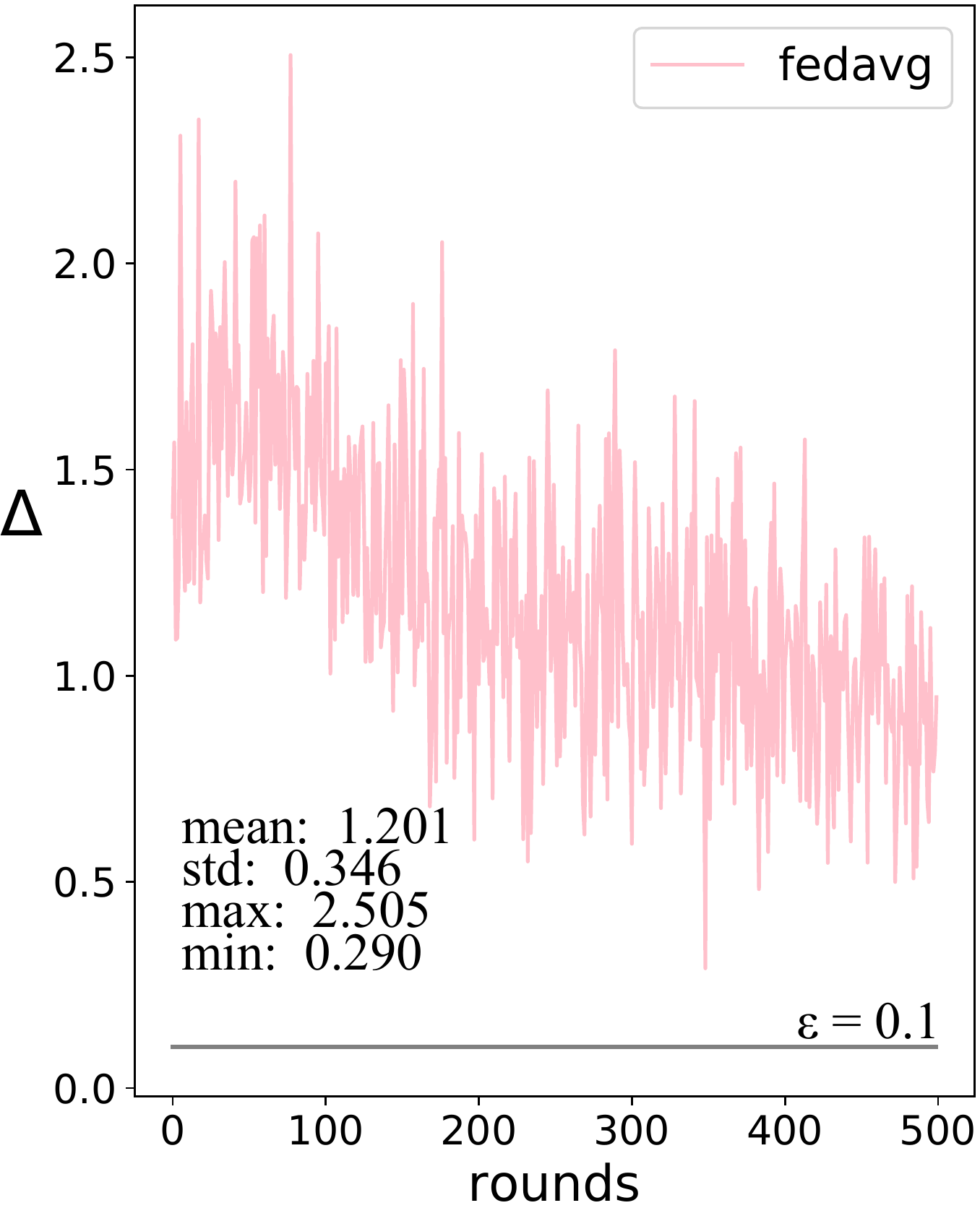}
    \caption{NFL using FedAvg.}
\end{subfigure}
\\
\vspace{3pt}
\begin{subfigure}[t]{0.49\columnwidth}
    \includegraphics[width=\columnwidth,height=0.75\columnwidth]{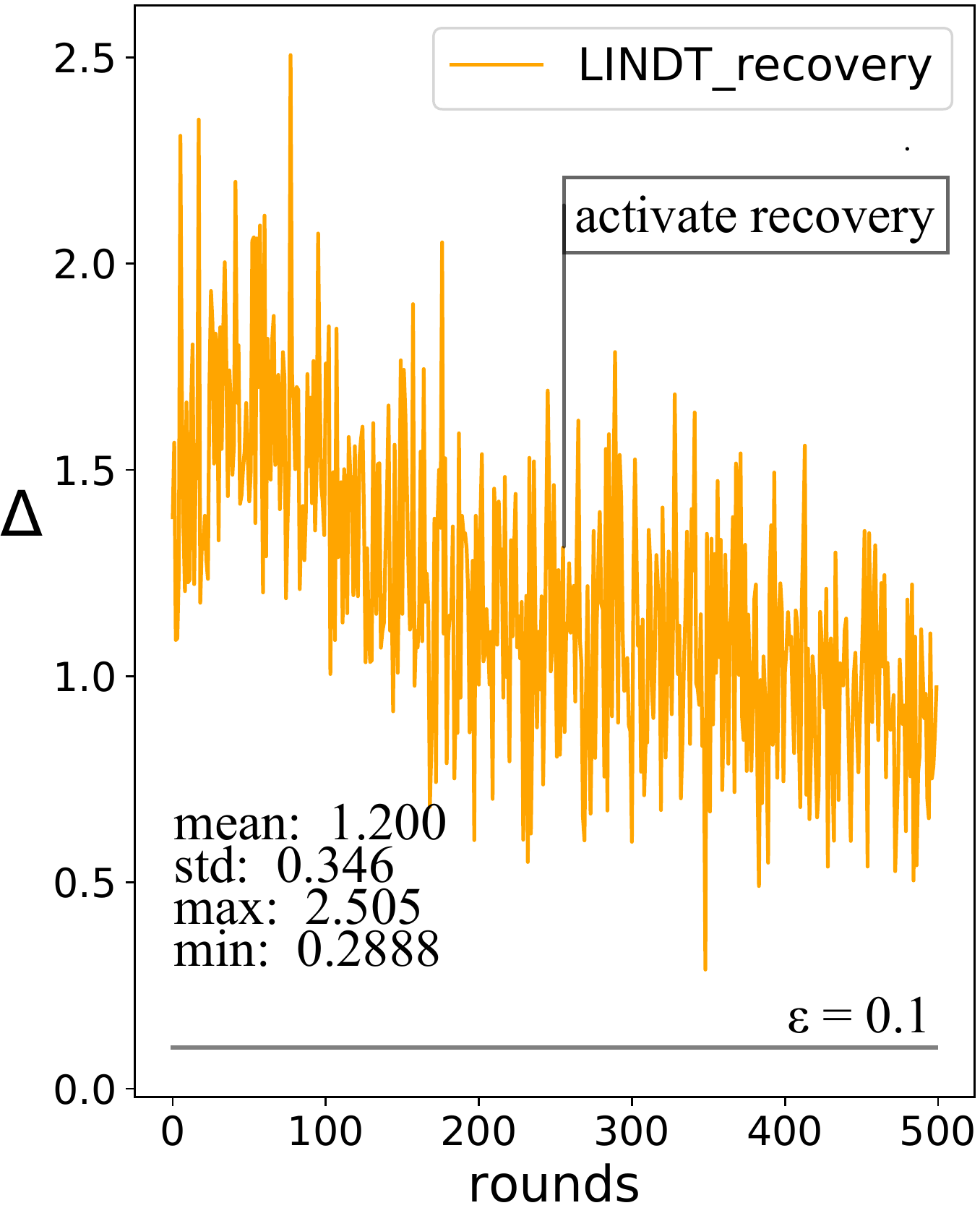}
    \caption{Recovery activated.}
\end{subfigure}
\hfil
\begin{subfigure}[t]{0.49\columnwidth}
    \includegraphics[width=\columnwidth,height=0.75\columnwidth]{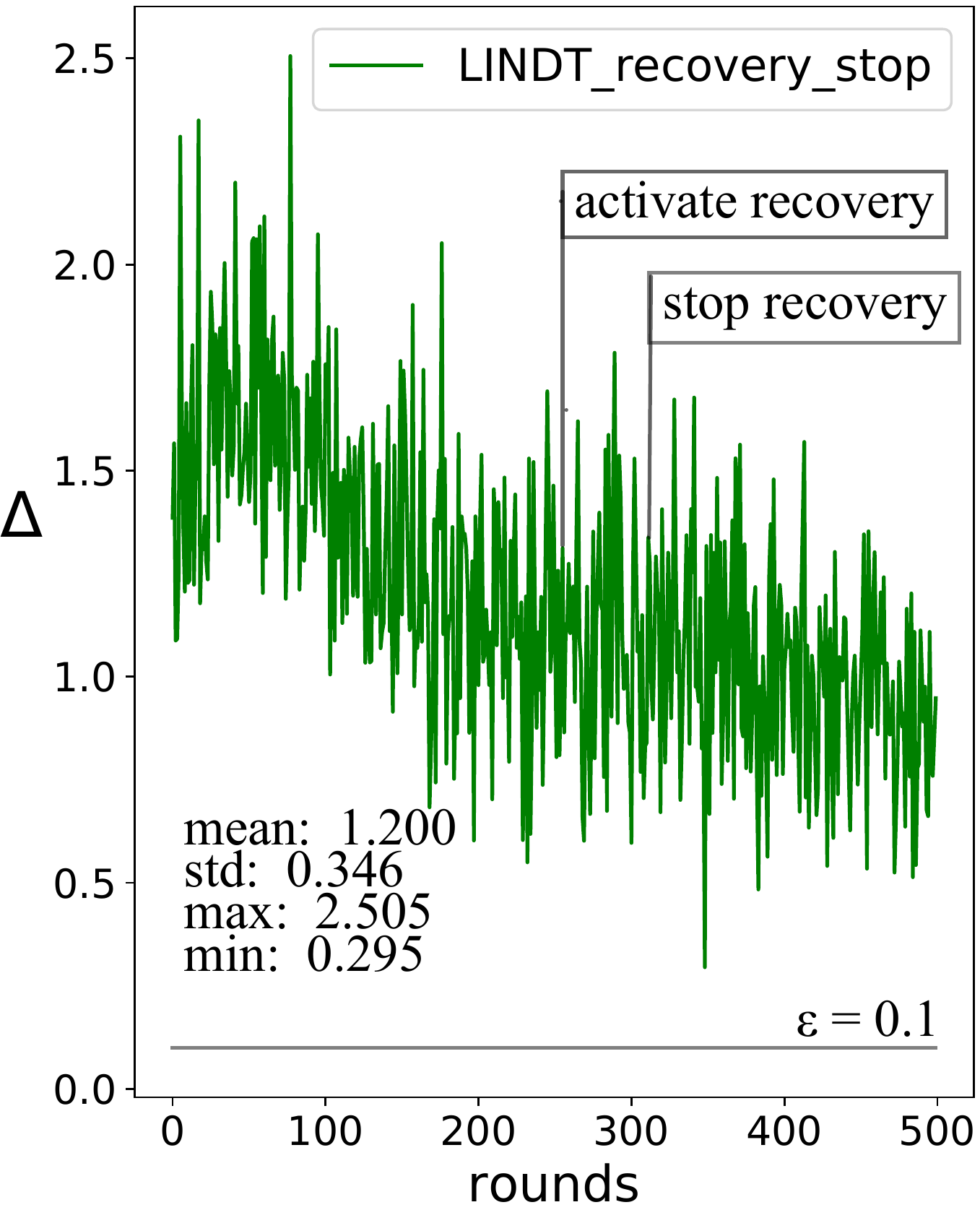}
    \caption{Recovery activated and later stopped.}
\end{subfigure}
\caption{Results of $\Delta$ in different FL processes.} 
\label{fig:results_of_Delta}
\end{figure}

As Fig.~\ref{fig:results_of_Delta} shows, the activation and stop of NFL recovery in LINDT does not cause visible changes in the value of $\Delta$ compared to that in FedAvg. Moreover, the statistics of different stopping strategies (shown in Fig.~\ref{fig:results_of_Delta}(b) and (c)) are both similar to FedAvg (Fig.~\ref{fig:results_of_Delta}(a)). These results reveal that, although a large $\Delta$ value indicates the likelihood of NFL, good performance in local accuracy (which is achieved by dual-model training in LINDT) does NOT ensure small $\Delta$. 
Therefore, $\Delta$ is NOT intended for measuring the quality of the ongoing federated learning, but only intended to measure \textit{the severity} of negative effects that are currently imposed on the federation. Only when most negative effects (as we discussed throughout the paper) are removed from the current federation environment, could the value of $\Delta$ converge to zero.


%

\subsection{When to Stop Recovery?}

\begin{figure}[h]
\centering
\includegraphics[width=0.6\columnwidth]{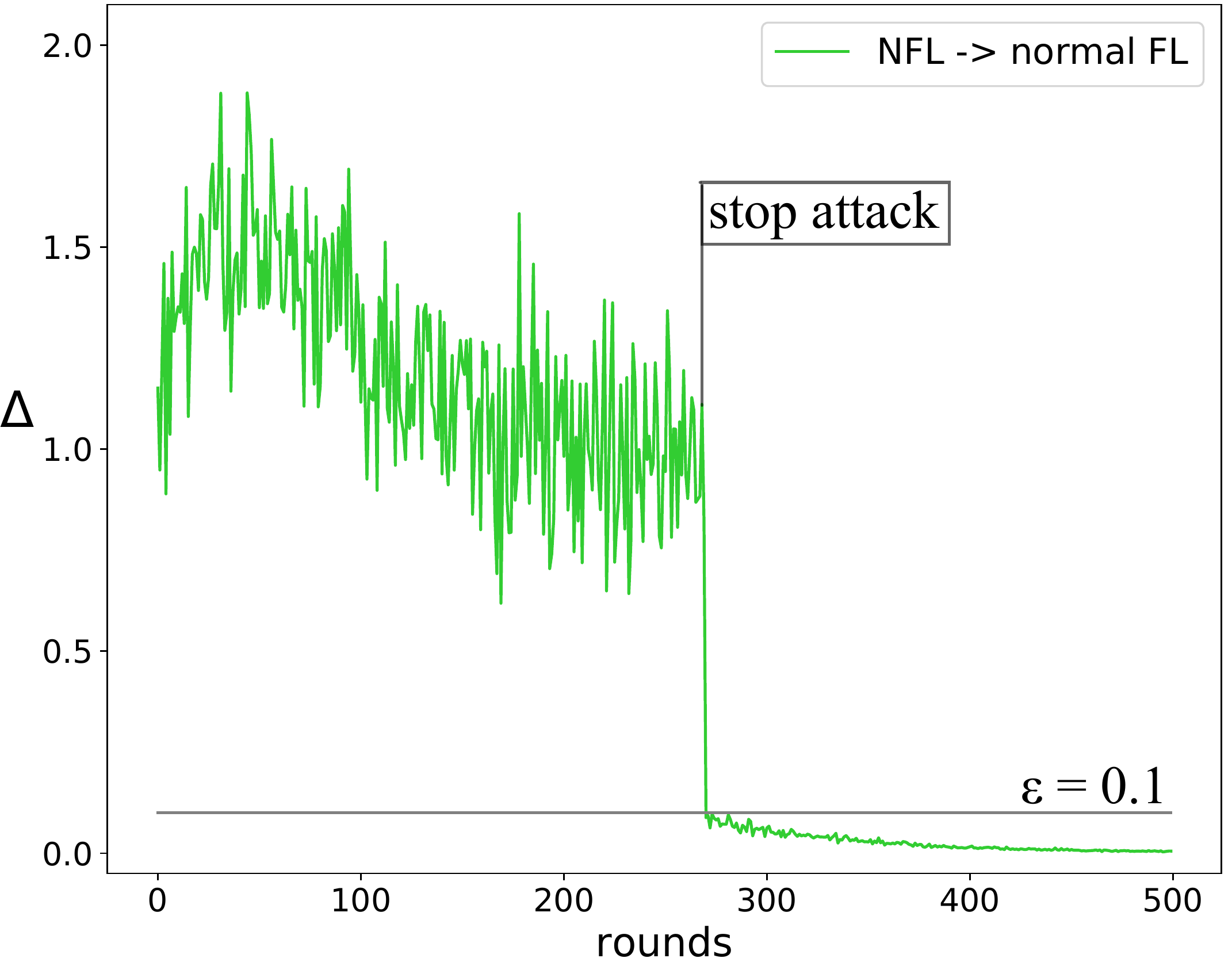}
\caption{Results of $\Delta$ where negative effects are removed at the flagged point. 
} 
\label{fig:w_div_NFL2normalFL}
\end{figure}

\begin{figure}[h]
\centering
\includegraphics[width=0.6\columnwidth]{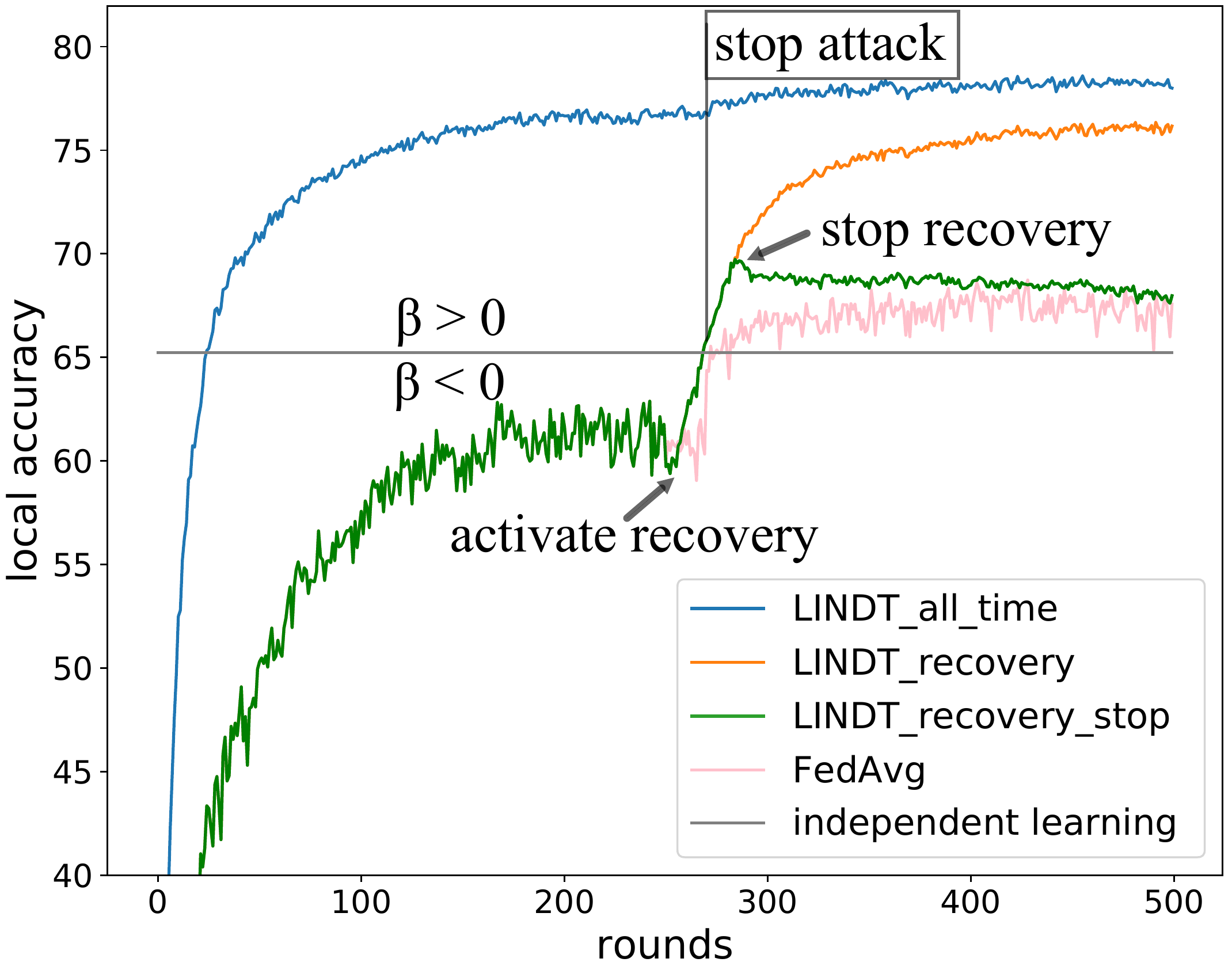}
\caption{Run-time performance of recovery where negative effects are removed at the flagged point. Recovery stops after the value of $\Delta$ is below $\epsilon$ in 10 consecutive rounds.
} 
\label{fig:recovery_stop2}
\end{figure}

The last experiment in the main paper shows that NFL recovery, once activated, cannot be stopped while the negative effects continue to exist, since, otherwise, the system may go back to NFL.
We would like to investigate recovery further by asking a question: Can we stop recovery when the system observes reduction in negative effects?

With this question in mind, we conduct an additional experiment on CIFAR by making amendments to the one in Section \ref{sec:detect_recovery}. Such amendments simulate a system (equipped with LINDT for NFL detection and recovery) which begins with NFL, and then at a later moment gets most of the negative effects removed. Since it is quite unreal to vary data distributions in run-time, we reallocate the data as \textit{non-IID(10)} (see Appendix \ref{C1} for details) to increase similarities in data distributions among the clients. In order to ensure the system enters NFL in the beginning, we retain the strength of attacks by setting 20\% of all clients to be attackers, and $\frac{K}{N}=0.1$ and $\sigma=0.001$ in differential privacy. The environment is designed so that the negative effects can be removed in a controlled way. 
Like the previous experiment, we still set $\epsilon=0.1$ and $r'=250$. 

As expected, LINDT detects NFL and activates recovery. Later we stop all attacks to remove most negative effects in run-time, and then monitor the performance under a new stopping strategy, i.e. the \textit{third strategy}, relying on $\Delta$. That is, the system stops dual-model training (and degrades to FedAvg) after the value of $\Delta$ is observed below $\epsilon$ in 10 consecutive rounds. For reference we also report the performance of the \textit{second strategy} as described in Section \ref{sec:detect_recovery} and that of FedAvg, as well as that of all-time LINDT.

Fig.~\ref{fig:w_div_NFL2normalFL} and Fig.~\ref{fig:recovery_stop2} show the experiment results. It can be seen that, as the recovery in LINDT proceeds, the local accuracy improves rapidly above the grey line, indicating that $\beta$ turns positive. When we remove the negative effects (at the flagged point), the value of $\Delta$ drops dramatically under $\epsilon$, in contrast to Fig. \ref{fig:results_of_Delta}(b,c) where $\Delta$ remains rough. Shortly after that, as the $\Delta$ value is kept under $\epsilon$ for 10 consecutive rounds, the system decides to stop dual-model training, and then the green curve extends almost horizontally without much 
downturn (as compared to Fig. \ref{fig:detect_recovery}(b)), and in the end, it terminates slightly above FedAvg (the pink curve) but far below the \textit{second strategy} (the orange one). The final state is reasonable, because the system already degrades to FedAvg after recovery stops. However, the short-term dual-model training has its lasting effects -- the performance appears more stable than FedAvg.

These results lead to three conclusions: First, the results strengthen the claim that $\Delta$ is indicative of the \textit{negative effects} rather than the \textit{learning performance} in the running system. Second, as an answer to the above question, it is relatively safe to stop recovery when  $\Delta$ reduces to insignificant values ($< \epsilon)$. Third, the local performance of LINDT is significantly better than FedAvg, both in time of NFL and normal FL.

\end{document}